\RequirePackage[T1]{fontenc}
\documentclass[letterpaper,10pt,conference]{ieeeconf}
\IEEEoverridecommandlockouts
\usepackage[utf8]{inputenc}
\usepackage{cite}
\usepackage{amsmath,amssymb}
\usepackage{graphicx,booktabs,tabularx,array,adjustbox}
\usepackage{capt-of}
\usepackage{xcolor,placeins}
\usepackage{algpseudocode}

\usepackage{needspace}
\makeatletter
\let\NAT@parse\@undefined
\makeatother
\definecolor{paperlink}{HTML}{1F4E79}
\usepackage[colorlinks=true,allcolors=paperlink]{hyperref}
\usepackage{xurl}
\newcommand{\vr}{VRQ}
\newcommand{\calD}{\mathcal{D}_{\mathrm{cal}}}
\newcommand{\clip}{\operatorname{clip}}
\newcommand{\roundop}{\operatorname{round}}
\newcommand{\nativepixel}{Pixel AUROC$_{96}$}
\newcounter{algorithm}
\newcolumntype{L}[1]{>{\raggedright\arraybackslash}p{#1}}

\graphicspath{{assets/}}

\newcommand{\VentBaseTime}{394.28}
\newcommand{\VentVTime}{407.12}
\newcommand{\VentSpeedup}{2.13}
\newcommand{\VentOverhead}{3.26}
\newcommand{\EntropyImageDelta}{0.198}
\newcommand{\EntropyPixelDelta}{0.074}

\newcommand{\EdgeAuthors}{%
Taewan Cho$^{1,2}$, Seungmin Lee$^{1}$, Jiho Seo$^{1,2}$, and Andrew Jaeyong Choi$^{1,*}$\\[0.8ex]
{\normalsize $^{1}$Intelligent Robotics and Autonomous System Control Laboratory (iRASC)}\\[0.3ex]
{\normalsize School of Computing, Gachon University, Republic of Korea}\\[0.3ex]
{\normalsize $^{2}$Plaid Labs Inc., Republic of Korea}\\[0.3ex]
{\small\texttt{\{taewan2002, magicmin2, ksdk6145, andrewjchoi\}@gachon.ac.kr}}\\[0.3ex]
{\small\texttt{\{technoking, swiftsjh\}@plaid.ai.kr}}%
\thanks{\protect\raggedright $^{*}$Corresponding author: Andrew Jaeyong Choi.}%
}

\newcommand{\PaperTitle}{EdgeZSAD: Value-Range Quantization for Deploying Zero-Shot Anomaly Detection on Edge Neural Processing Units}
\hypersetup{pdftitle={\PaperTitle},
  pdfauthor={Taewan Cho, Seungmin Lee, Jiho Seo, and Andrew Jaeyong Choi},
  pdfsubject={VRQ-PTQ: Value-range diagnosis, source-only correction, and QNN/HTP validation},
  pdfkeywords={VRQ-PTQ, VRQ, Value-Range Quantization, MinMax, zero-shot anomaly detection, post-training quantization}}

\begin{document}
\title{\LARGE\bf \PaperTitle}
\author{\EdgeAuthors}
\maketitle
\raggedbottom

\setlength{\emergencystretch}{3em}
\setlength{\textfloatsep}{1\baselineskip plus .2\baselineskip minus .2\baselineskip}
\setlength{\dbltextfloatsep}{1\baselineskip plus .2\baselineskip minus .2\baselineskip}
\setcounter{topnumber}{5}
\setcounter{dbltopnumber}{4}
\renewcommand{\topfraction}{.9}
\renewcommand{\dbltopfraction}{.9}
\renewcommand{\textfraction}{.05}
\renewcommand{\floatpagefraction}{.65}
\renewcommand{\dblfloatpagefraction}{.65}

\begin{abstract}
\normalfont\small
Quantizing zero-shot industrial anomaly detectors for edge neural processing units can degrade defect detection and localization. We present EdgeZSAD (Edge Zero-Shot Anomaly Detection), a deployment framework whose methodological contribution is Value-Range Quantization (VRQ). We diagnose how attention Value tensors inherit overly wide ranges from fused query--key--value projections and assign independent ranges calibrated on source data, without retraining or changing quantized weights. The engineering application is native deployment of a foundation-model detector for industrial inspection, evaluated on three public benchmarks and Qualcomm edge devices. Under minimum--maximum and mean-squared-error calibration, VRQ improves image-level discrimination and per-region overlap. In the primary calibration set, it recovers 90--98\% of the loss in combined image and pixel discrimination relative to half precision, with a small increase in latency over baseline quantization. Matched tensor overrides reproduce the recovery without adding graph nodes, identifying independent Value-range assignment as the effective intervention in the evaluated runtime family.

\end{abstract}
\section{Introduction}
In high-mix, low-volume manufacturing, frequent product changes can make automated visual inspection costly to maintain. Defect examples may be scarce, while even normal-only anomaly detectors can require product-specific training data or reference sets. Preparing these resources and updating models creates recurring work. For small and medium-sized manufacturers, limited specialist staff, investment capacity and digital infrastructure can constrain adoption~\cite{oecd_sme_ai}. Inspection systems therefore need to reduce both product-specific preparation and the resources required for routine operation.

Zero-shot anomaly detection transfers a detector trained on auxiliary data to unseen inspection targets, reducing product-specific training requirements. Deploying a foundation-model detector at multiple inspection stations also requires attention to memory, equipment cost, power and cooling. Edge neural processing units (NPUs) offer a deployment route, provided that model conversion preserves both defect detection and localization.

EdgeZSAD investigates how much of a pretrained zero-shot detector's quality can be retained under quantization for native NPU execution. We use the released AnomalyVFM detector with the Reduce All Domains Into One (RADIO) backbone~\cite{anomalyvfm,radio}. Post-training quantization (PTQ) reduces precision without retraining, but its effectiveness depends on where a calibrated range is assigned. An attention operator may inherit an unnecessarily wide grid for its Value input, changing anomaly rankings and threshold decisions.

In the evaluated Qualcomm Neural Network (QNN) stack and Hexagon Tensor Processor (HTP) backend, Value inputs to probability--value (PV) multiplication inherit the fused query--key--value (QKV) grid through layout operators. In all 24 blocks, this grid is wider than source-observed Value. Native range, precision and location controls isolate the effect of assigning Value its own grid where it is consumed.

\begin{figure*}[t]
\centering
\includegraphics[width=\textwidth]{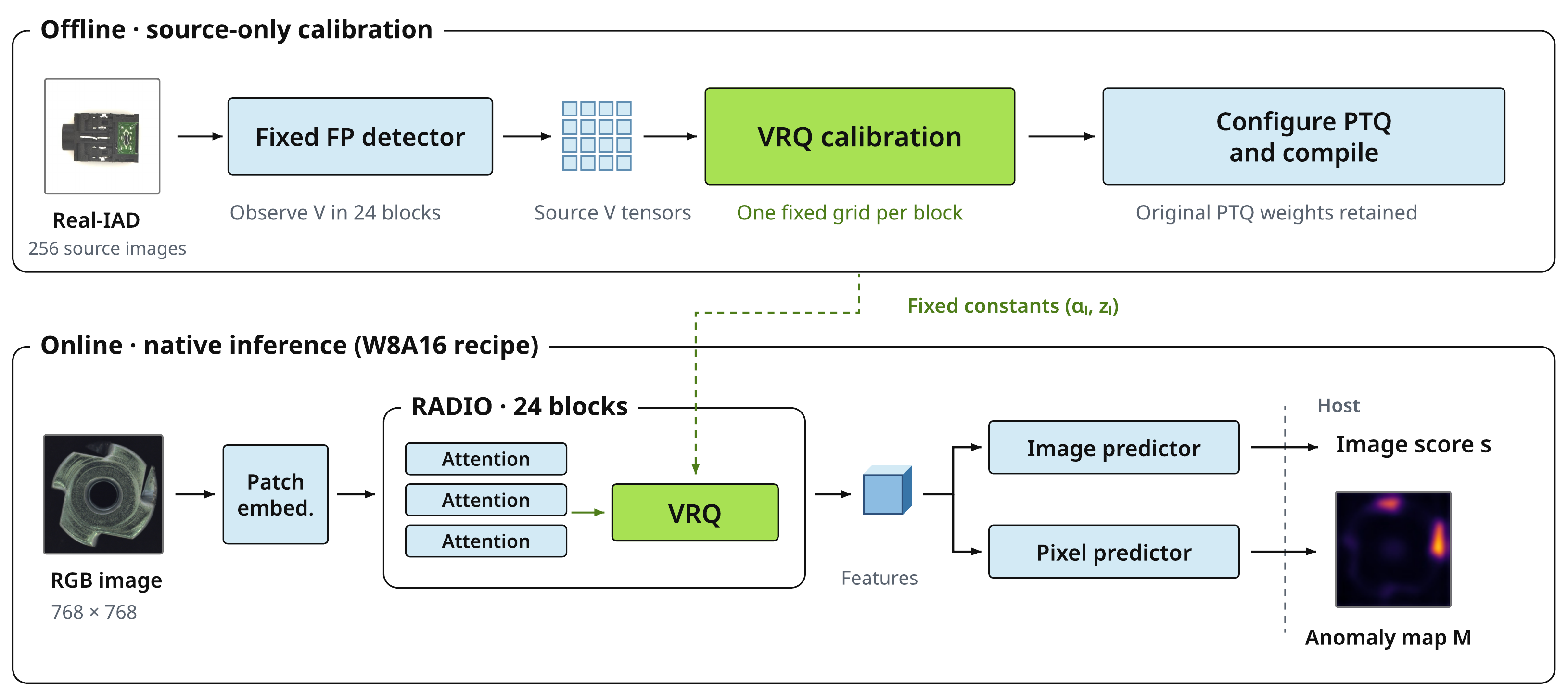}
\caption{VRQ overview: source-only Value calibration and native inference. FP denotes floating point. Fixed calibration constants configure the Value grid before compilation; runtime uses the existing detector and predictors. The metal-nut map is an Entropy PTQ+VRQ output illustrating the interface. MinMax PTQ+VRQ is the representative deployment configuration evaluated in the results.}
\label{fig:method}
\end{figure*}

\emph{Value-Range Quantization (VRQ)} assigns each block a fixed affine grid observed from its complete source Value tensor. It preserves existing PTQ weights and ordinary activation encodings, requires no retraining or target calibration, and supports added quantize--dequantize (QDQ) pairs or a tensor override. We denote the combined deployment \emph{VRQ-PTQ} (Figure~\ref{fig:method}).

Matched PTQ/VRQ pairs evaluate 4,971 images from three datasets under two Real-IAD source selections: set A on VENTUNO Q and set B on IQ9075 (Section~\ref{sec:setup}). Both minimum--maximum (MinMax) and mean-squared-error (MSE) calibration improve in the separate discrimination and region-overlap means. MinMax PTQ+VRQ is our representative configuration; within-observer recovery is the methodological result.

The methodological contribution is diagnosed range assignment: native controls locate the affected operand, and the matched override and added-QDQ implementations separate the correction from node insertion. The engineering contribution is native quality--cost validation across detection, localization, source-fixed decisions and QNN/HTP boards, separating PTQ's deployment savings from VRQ's incremental recovery and cost.

\section{Related Work}
\paragraph{Industrial anomaly inspection}
Industrial methods combine density estimation, metric learning and teacher--student representations to detect deviations from normal products. Normalizing flows have been combined with multiscale ensembles~\cite{eaai_inai}; ProxyCore combines proxy metric learning with coresets~\cite{eaai_proxycore}; dual-branch reverse distillation supports screw inspection~\cite{eaai_screw}; and normal-sample adjustment has been studied for stamped-part cracks~\cite{eaai_stamping}. These methods motivate evaluating both image decisions and defect localization. Our focus is preserving an existing detector when converting it to native low-precision execution.

\paragraph{Foundation-model zero-shot detection}
WinCLIP uses prompt ensembles and window features~\cite{winclip}, AnomalyCLIP learns object-agnostic prompts from auxiliary data~\cite{anomalyclip}, and MuSc scores unlabeled test images jointly~\cite{musc}. Prompt adaptation, modality alignment and expert fusion have also been explored in industrial inspection~\cite{eaai_ppta,eaai_modal,eaai_emain}. AnomalyVFM instead adapts vision foundation models for anomaly detection using RADIO representations~\cite{anomalyvfm,radio}. We use its released detector with static per-image inference; target images fit neither calibration constants nor a memory bank. The detector comparison in Appendix~\ref{app:detector_context} provides application context, while the within-detector PTQ pairs isolate the contribution of VRQ.

\paragraph{Quantization of vision transformers}
Reconstruction methods optimize weight rounding, block outputs or prediction differences~\cite{adaround,brecq,qdrop,pdquant}; subsequent work refines curvature approximations and reconstruction objectives~\cite{aphq,fimaq,lsvit}. PTQ4ViT and FQ-ViT design quantizers for attention and nonlinear activation distributions~\cite{ptq4vit,fqvit}. RepQ-ViT and RepQuant use scale reparameterization~\cite{repq,repquant}, while SmoothQuant, NoisyQuant and instance-aware grouping redistribute quantization difficulty or adjust granularity~\cite{smoothquant,noisyquant,igqvit}. Dense-prediction methods address Key/attention distributions and hardware-compatible formats~\cite{ptq4sam,ahcptq}. Q-HyViT uses bridge-block reconstruction to search quantization choices in hybrid transformers~\cite{qhyvit}. These approaches change reconstruction, quantizer format or granularity. VRQ addresses the range assigned to a particular consumed attention operand, retaining tensorwise affine activation grids and existing quantized weights.

\paragraph{Hardware-aware deployment}
Affine integer quantization enables efficient inference when its operators and encodings map to the target backend~\cite{jacob2018}. I-ViT targets integer transformer execution~\cite{ivit}, joint pruning/quantization reduces edge-model cost~\cite{eaai_sspq}, and edge deployment studies emphasize the software and hardware stack~\cite{eaai_edgestack}. Official tensor overrides provide a practical mechanism for independent range assignment~\cite{ortoverride}. We combine that mechanism with a diagnosis of the affected attention boundary and native quality/cost measurements. The override control is therefore a test of the VRQ principle alongside the added-QDQ realization.

\section{Problem Diagnosis}
\subsection{Detector contract and quantization setting}
The fixed AnomalyVFM/RADIO detector maps a $1\times3\times768\times768$ input to an image logit $u(x)$ and $1\times1\times96\times96$ pixel logits $Z(x)$. The image score and native map are
\begin{align}
s(x)&=\sigma\bigl(u(x)\bigr),\label{eq:score}\\
M(x)&=\operatorname{AvgPool}_{5\times5}\bigl(\sigma(Z(x))\bigr),\label{eq:map}
\end{align}
where $\sigma$ is the logistic sigmoid; pooling uses stride 1 and zero padding of 2. The primary pixel metric resizes this prediction to the original ground-truth mask, as specified in Section~\ref{sec:setup}.

\label{sec:w8a16}
W8A16 denotes 8-bit weight and 16-bit integer activation encodings. It does not denote half-precision floating-point activations. The graph stores unsigned integer weights with per-output-channel scales and scalar activation scales. A native first-block probe exposes a Value-path conversion from unsigned 16-bit to unsigned 8-bit fixed point before PV. We therefore distinguish the exported QDQ grid from the compiler's internal types and measure native output error.

\subsection{Range propagation at the Value boundary}
\label{sec:diagnosis}
In each block, Value's QDQ pair references the fused QKV output scale and zero point. Open Neural Network Exchange (ONNX) Runtime propagates these constants through Reshape, Split, Transpose and Squeeze operations; 22 quantize/dequantize nodes per block share one scale initializer. Query and Key instead pass through scaling operations and receive their own grids. In the first block, Query's scale matches the source estimate and Key differs by one float32 unit in the last place with the same zero point. Value occupies only $1/5.89$ of the inherited range.

\begin{figure*}[t]
\centering
\includegraphics[width=\textwidth]{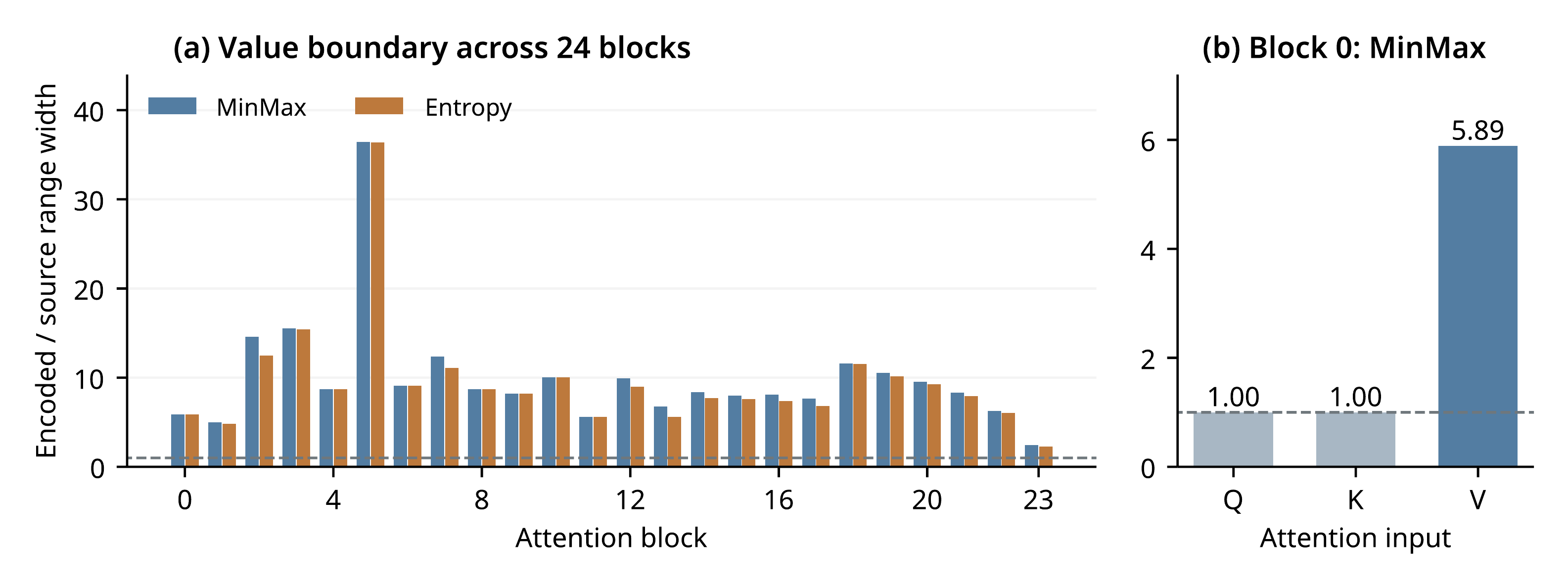}
\caption{ONNX range diagnosis. (a) MinMax (blue) and Entropy (orange) Value-grid widths relative to FP source-observed widths at all 24 PV inputs. All exceed the equal-width line (medians 8.53 and 8.45). (b) The separate MinMax block-0 audit finds matching Q/K ranges and a 5.89-fold wider V range, inherited from the fused QKV projection. Ratios describe exported encodings; native behavior is tested separately.}
\label{fig:range_widths}
\end{figure*}

Figure~\ref{fig:range_widths} extends this audit to all 24 blocks. The encoded-to-observed width ratios range from 2.42 to 36.43 for MinMax, with median 8.53, and from 2.25 to 36.41 for Entropy, with median 8.45. Every Value input carries a wider projection grid. This identifies a systematic range-propagation effect in the exported attention layout and motivates correcting the boundary rather than changing all attention tensors.

\subsection{Native range and location controls}
The first-block probe supplies identical floating-point (FP) prefix outputs to all conditions on 16 source-reserve images. Relative root-mean-square (RMS) error is $\|Y_{\mathrm{test}}-Y_{\mathrm{FP}}\|_2/\|Y_{\mathrm{FP}}\|_2$. Table~\ref{tab:mechanism} crosses range and bit-width. Adding QDQ on the existing grid produces the same native output as the baseline. Keeping 16-bit encoding but assigning the source-observed Value range reduces native error from 3.5451\% to 1.2646\%, a 64.33\% reduction, and improves all 16 inputs. Changing only Value's bit-width under the wide range does not reduce the error.
\begin{table}[!tbp]
\centering\small
\caption{Range $\times$ bit-width control: isolated first-block relative RMS (\%, lower is better) on 16 source validation images. FP--CPU measures QDQ reference error; FP--NPU measures native error against the same FP reference.}
\label{tab:mechanism}
\setlength{\tabcolsep}{3pt}
\begin{adjustbox}{max width=\linewidth}
\begin{tabular}{lrr}
\toprule
\textbf{V condition at PV} & \textbf{FP--CPU} & \textbf{FP--NPU}\\
\midrule
Anchor (baseline PTQ) & 0.9208 & 3.5451 \\
Wide range / V16 & 0.9208 & 3.5451 \\
V-specific range / V16 & 0.9207 & 1.2646 \\
Wide range / V8 & 3.1598 & 3.5451 \\
V-specific range / V8 & 1.0185 & 1.2674 \\
 
\bottomrule
\end{tabular}
\end{adjustbox}
\end{table}

The floating-point-to-central-processing-unit (CPU) QDQ error changes only from 0.9208\% to 0.9207\% when Value keeps its 16-bit encoding, because the observed Value span still covers about 11,000 of the 65,535 levels of the inherited grid. The native errors instead follow the CPU rows that encode Value with 8 bits---3.1598\% under the wide range and 1.0185\% under the Value-specific range---and are almost unaffected by the declared Value bit-width: 3.5451\% for both wide-range rows, and 1.2646\% versus 1.2674\% for the Value-specific rows. This pattern agrees with the observed fixed-point conversion. If PV consumes Value at 8 bits, the inherited range leaves the observed span about 43 of 255 levels in the first block, so the assigned range, rather than the exported bit-width, sets the effective resolution. Exported 16-bit QDQ simulation alone therefore misses the tested backend's sensitivity to the Value range. Representative native probes in blocks 12 and 23 reduce errors from 2.0581\% to 1.1204\% and from 1.2745\% to 0.9833\%, respectively.

Table~\ref{tab:qkv} isolates intervention location. Query-only, Key-only, Query+Key and same-grid controls match the baseline output. Value-only correction improves the result, and Query+Key+Value matches Value-only. Because Query and Key already have matching ranges, these controls identify the misplaced Value range rather than a general lack of Query/Key sensitivity.
\begin{table}[!tbp]
\centering\small
\caption{First-block intervention-location control: native relative RMS against FP and reduction from baseline. The 16 source inputs share a fixed FP prefix. Original Q/K ranges already match or nearly match source observations.}
\label{tab:qkv}
\setlength{\tabcolsep}{3pt}
\begin{adjustbox}{max width=\linewidth}
\begin{tabular}{lrr}
\toprule
\textbf{Added calibration} & \textbf{Error (\%)} & \textbf{Reduction (\%)}\\
\midrule
Baseline PTQ & 3.5451 & 0.00 \\
Q same-grid & 3.5451 & 0.00 \\
K same-grid & 3.5451 & 0.00 \\
Q-only range & 3.5451 & 0.00 \\
K-only range & 3.5451 & 0.00 \\
V-only range & 1.2646 & 64.33 \\
Q+K range & 3.5451 & 0.00 \\
Q+K+V range & 1.2646 & 64.33 \\
 
\bottomrule
\end{tabular}
\end{adjustbox}
\end{table}

\section{Value-Range Quantization (VRQ)}
\label{sec:vrq_ptq}
VRQ assigns an independently source-calibrated grid to each diagnosed Value input to PV; VRQ-PTQ denotes a baseline PTQ graph combined with VRQ. The baseline observer estimates ordinary activation ranges; the VRQ component assigns the Value boundary's range while preserving the baseline weights and Q/K paths. MinMax PTQ+VRQ is the representative configuration: both stages use MinMax observations but operate on different tensors. Entropy PTQ+VRQ and MSE PTQ+VRQ retain the same source-MinMax Value rule while changing the ordinary activation observer. Each block uses one scalar Value scale and zero point, without head-, token-, or channel-wise activation grids. Algorithm~\ref{alg:vrange} specifies the added-QDQ realization; tensor-override controls test whether the additional nodes are necessary. Native re-evaluation checks the result after compilation.

\begin{figure*}[t]
\centering
\includegraphics[width=\textwidth]{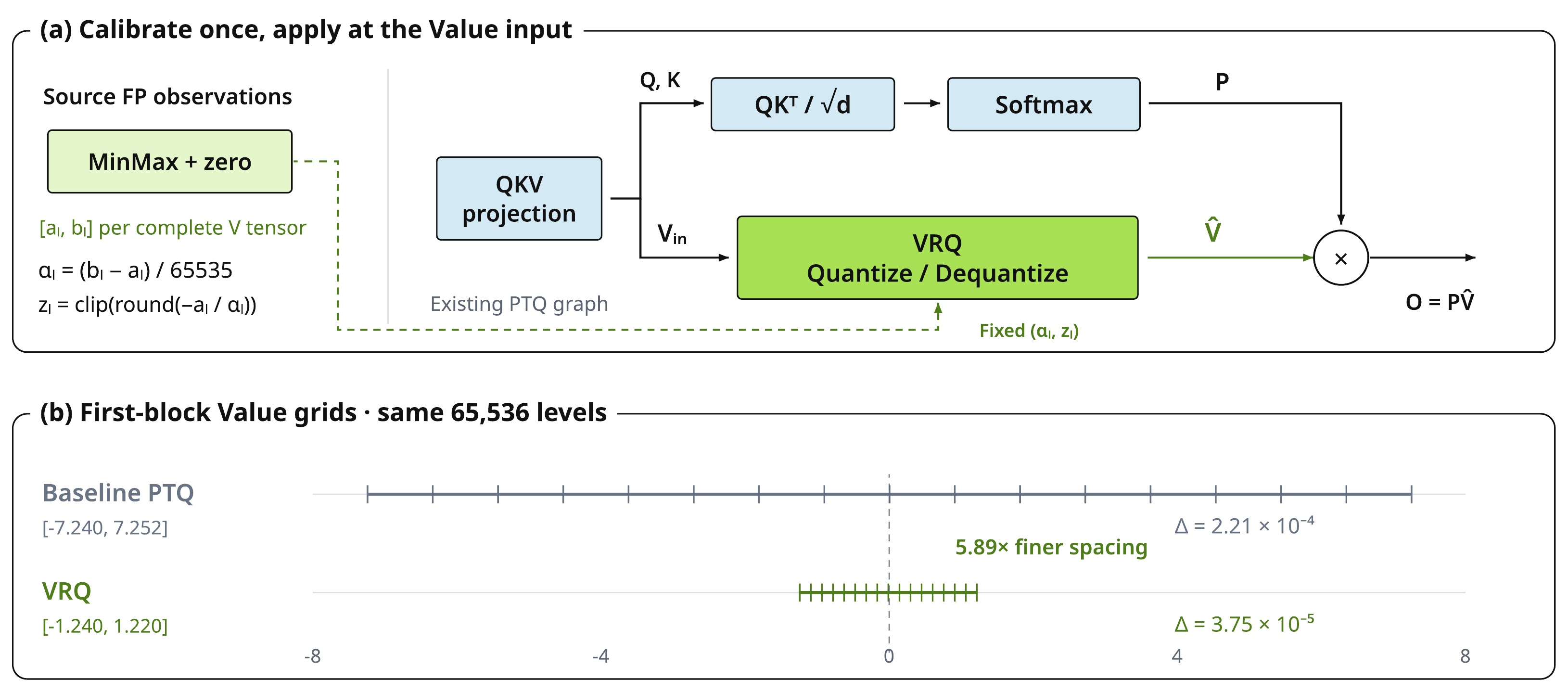}
\caption{PTQ and \vr{} at the ONNX PV boundary. (a) \vr{} adds a source-calibrated UINT16 grid; dashed arrows supply fixed constants. Existing weights and QK/Softmax remain unchanged. (b) First-block grids on a shared axis (MinMax and Entropy coincide; the PTQ grid is the fused QKV projection's): the same 65,536 levels give a $5.89\times$ smaller step, and the ratio is the same at any fixed level count. Ticks are subsampled; compiled precision is evaluated separately (Table~\ref{tab:mechanism}).}
\label{fig:vrange_algorithm}
\end{figure*}

\subsection{Attention and the intervention boundary}
For query, key, and value tensors $Q,K,V$ in one attention head,
\begin{equation}
P=\operatorname{softmax}\!\left(\frac{QK^{\mathsf T}}{\sqrt{d}}\right),
\qquad O=PV. \label{eq:attention}
\end{equation}
Here $d$ is the head's query/key dimension. Because $P$ and $V$ differ in role and distribution, we separately calibrate each block's complete V tensor, without per-token/head/channel scales or QK/softmax redesign.

For block $l$, $V_l^{\mathrm{FP}}$ denotes FP source observations, while $V_l^{\mathrm{in}}$ is the actual PV-boundary input from the PTQ graph. The former fixes the grid; the latter is quantized at runtime and retains upstream quantization error.

\subsection{Source ranges and fixed affine grids}
For $l\in\{0,\ldots,23\}$, index $i$ spans all heads, tokens, and channels of the complete Value tensor. Using source set $\calD$, we define a range containing zero:
\begin{align}
a_l &= \min\!\left(0,\min_{x\in\calD,\,i}V^{\mathrm{FP}}_{l,i}(x)\right),\\
b_l &= \max\!\left(0,\max_{x\in\calD,\,i}V^{\mathrm{FP}}_{l,i}(x)\right). \label{eq:range}
\end{align}
All observed blocks satisfied $b_l>a_l$. With $K=2^{16}-1=65535$, we compute
\begin{align}
\alpha_l&=(b_l-a_l)/K,\\
z_l&=\clip\!\left(\roundop(-a_l/\alpha_l),0,K\right). \label{eq:grid}
\end{align}
For a scalar $v$ from runtime input $V_l^{\mathrm{in}}$, the transformation is
\begin{align}
q_l(v)&=\clip\!\left(\roundop(v/\alpha_l)+z_l,0,K\right),\\
\widehat v_l&=\alpha_l\bigl(q_l(v)-z_l\bigr). \label{eq:qdq}
\end{align}
Rounding uses nearest-even ties. We convert the scale to float32 before computing the zero point with \texttt{np.rint}, then store it as uint16. The fixed grid transforms $V_l^{\mathrm{in}}$ elementwise into $\widehat V_l$ (Figure~\ref{fig:vrange_algorithm}a). Standard affine QDQ~\cite{jacob2018} lets us express the intervention using existing operators, without custom kernels in the evaluated QNN path.

\subsection{Graph integration and deployment}
Algorithm~\ref{alg:vrange} specifies the added-QDQ path. The FP CPU observer disables graph optimization. At each PV MatMul, a QuantizeLinear--DequantizeLinear pair is inserted on the second input edge after the head layout: 24 pairs, scales, and zero points. Alternatively, excluding the 24 Value Squeeze nodes from QDQ parameter sharing allows ordinary independent calibration; official tensor overrides~\cite{ortoverride} can assign the same fixed ranges without the extra pairs. Exclusion changes quantization policy, not the detector operation. Section~\ref{sec:independent_setup} compares these implementations under native execution checks.

Calibration uses no label-dependent loss, backpropagation or target-specific optimization. Source selection is specified in Section~\ref{sec:setup}.

\begin{figure}[!tbp]
\centering
\begin{minipage}{.97\linewidth}
\hrule\vspace{5pt}
\refstepcounter{algorithm}\textbf{Algorithm \thealgorithm. Source-only VRQ calibration}\label{alg:vrange}
\smallskip

\footnotesize
\algrenewcommand{\algorithmicrequire}{\textbf{Input:}}
\algrenewcommand{\algorithmicensure}{\textbf{Output:}}
\algrenewcommand{\algorithmicindent}{1em}
\begin{algorithmic}[1]
\Require Fixed FP detector $f_{\mathrm{FP}}$, PTQ graph $G_0$, source set $\calD$
\Ensure Graph $G$ with fixed Value-input grids
\State $G\gets\Call{Copy}{G_0}$; $K\gets 2^{16}-1$
\State $a_l\gets0$, $b_l\gets0$ for $l=0,\ldots,23$ \Comment{include zero}
\For{each $x\in\calD$}
  \State $\{V_l^{\mathrm{FP}}(x)\}_{l=0}^{23}\gets\Call{ObserveValues}{f_{\mathrm{FP}},x}$
  \For{$l=0,\ldots,23$}
    \State $a_l\gets\min(a_l,\min_i V_{l,i}^{\mathrm{FP}}(x))$
    \State $b_l\gets\max(b_l,\max_i V_{l,i}^{\mathrm{FP}}(x))$
  \EndFor
\EndFor
\For{$l=0,\ldots,23$}
  \State \textbf{assert} $b_l>a_l$
  \State $\alpha_l\gets\operatorname{float32}((b_l-a_l)/K)$
  \State $z_l\gets\operatorname{uint16}(\clip(\roundop(-a_l/\alpha_l),0,K))$
  \State $G\gets\Call{InsertValueQDQ}{G,l,\alpha_l,z_l}$
\EndFor
\State \Call{VerifyUnchanged}{$G_0,G$}
\State \Return $G$
\end{algorithmic}
\smallskip
\textsc{ObserveValues} uses the FP CPU observer without graph optimization. \textsc{InsertValueQDQ} fixes the UINT16 grid at PV's second input. \textsc{VerifyUnchanged} checks the original initializers and untargeted edges. Compilation and source-execution checks precede target evaluation.
\vspace{4pt}\hrule
\end{minipage}
\end{figure}

\subsection{Value rounding error and applicability}
Let $V_{ik}$ and $\widehat V_{ik}$ denote elements of $V_l^{\mathrm{in}}$ and $\widehat V_l$ in one head. Without clipping and with fixed $P$, each element's additional QDQ rounding error is at most $\alpha_l/2$. Since $P$ has nonnegative rows summing to one,
\begin{equation}
\left|\sum_i P_{ji}(\widehat V_{ik}-V_{ik})\right|
\leq\sum_iP_{ji}\frac{\alpha_l}{2}
=\frac{\alpha_l}{2}. \label{eq:bound}
\end{equation}
The convex combination in Equation~\eqref{eq:bound} keeps this local bound independent of token count. At fixed bit-width, narrowing the span reduces the bound proportionally---by $5.89\times$ for the first-block example. The ratio does not depend on the level count: Equation~\eqref{eq:grid} states the exported 16-bit grid, whereas if PV consumes Value at 8 bits, as the native first-block probe indicates (Section~\ref{sec:w8a16}, Table~\ref{tab:mechanism}), the same narrowing reduces the step from approximately 0.057 to 0.0096 rather than from $2.21\times10^{-4}$ to $3.75\times10^{-5}$. The bound concerns only added rounding of the runtime Value input; clipping, upstream error, quantized $P$, and compiler transformations lie outside its assumptions. Section~\ref{sec:results} tests the resulting end-to-end effect through native evaluation.

\section{Experimental Setup}
\label{sec:setup}
\subsection{Detector, source data and matched comparisons}
The released AnomalyVFM/RADIO detector has 24 attention blocks, embedding width 1024 and patch size 16~\cite{anomalyvfm,radio}. RGB inputs are bilinearly resized to $768\times768$, converted to float32 in channel--height--width order and divided by 255. Batch size is one; predictors are unchanged. Appendix~\ref{app:deployment} identifies the checkpoint, exporter and runtimes.

Each set contains 256 Real-IAD camera-1 images from 19 categories: 128 normal and 128 anomalous~\cite{realiad}. Labels balance selection, without a label-dependent calibration loss. A/B use distinct ordered source selections and fixed exports, evaluated on VENTUNO Q/IQ9075 respectively. Within each observer/VRQ pair, source/target inputs, weight encodings, ordinary activation ranges and postprocessing are identical; cross-set comparisons confound several factors. Source-reserve images are product-group disjoint from calibration, and file/decoded-image checks exclude source--target duplicates.

References use graphics processing unit (GPU) FP32 and native HTP FP16 execution (32-/16-bit floating point). Baselines use unsigned 8-bit weights with per-output-channel scales and unsigned 16-bit activations with scalar scales. Ordinary observers are MinMax, Entropy or MSE; the MSE search is described in Appendix~\ref{app:mse_implementation}. All VRQ variants use source-MinMax Value ranges. Added-QDQ models preserve all 1,986 baseline initializers, including quantized weights.

\label{sec:independent_setup}
The tensor-override controls exclude 24 Value Squeeze nodes from parameter sharing and use ONNX Runtime tensor overrides~\cite{ortoverride} to assign the same Value-MinMax grids without adding QDQ nodes, retaining baseline weights and ordinary encodings. Tables label this implementation as VRQ (override); VRQ alone denotes added QDQ. An additional Entropy control independently observes Value with Entropy. Fixed-grid AdaRound optimizes rounding on the same source images, retaining activation grids and topology~\cite{adaround}.

\subsection{One primary target and metric protocol}
Every full-dataset condition evaluates the same 4,971 targets: 1,725 MVTec AD images in 15 categories~\cite{mvtec}, 2,162 VisA images in 12 categories~\cite{visa}, and 1,084 images from the public MVTec AD 2 split in 8 categories~\cite{mvtecad2}. We abbreviate the latter dataset as AD2 public. No target images fit quantization ranges, model weights or decision thresholds.

Image area under the receiver operating characteristic curve (AUROC)~\cite{fawcett} uses the image-head score in Equation~\eqref{eq:score}. Pixel AUROC uses Equation~\eqref{eq:map}, bilinearly upsampled to the original mask resolution with \texttt{align\_corners=False}. Ground truth retains its original resolution, values greater than zero denote anomalies, and normal images are included. We compute each metric per category and take an equal-category mean. Values are reported as $\times100$ and differences as percentage points (pp). Appendix~\ref{app:complete_native} describes native-grid metrics.

Area under the per-region overlap curve (AUPRO) measures defect-region coverage up to false-positive-rate (FPR) limits 0.05 and 0.30. Regions are 8-connected components of the original mask. Each region receives equal weight within its image; non-anomalous pixels, including normal images, enter the FPR denominator. Exact integration over distinct score levels and linear interpolation at the FPR limit follow the retained evaluation implementation and cross-checks~\cite{mvteceval}.

\emph{3-set AUROC Mean} gives Image and Pixel AUROC equal weight within each dataset, then weights the three datasets equally. \emph{3-set AUPRO Mean} gives the two FPR limits equal weight within each dataset and again weights datasets equally. These are separate discrimination and region-overlap summaries. Set A's displayed aggregate means use the original uniformly rounded table cells; set B's means use unrounded scores.

\subsection{Uncertainty and source-fixed decisions}
Dataset-level Image intervals use 2,000 paired bootstrap resamples within fixed categories: images for MVTec/VisA and physical scenes with all lighting views for AD2. Separate Image/Pixel aggregate intervals use paired cluster jackknife on six saved conditions, retaining Image--Pixel covariance in each deletion. 

Each condition's operating threshold uses the same rule on 608 additional Real-IAD normal images, without target selection. Category-macro false-negative rate (FNR) and FPR expose the decision trade-off alongside threshold-independent AUROC/AUPRO.

\subsection{Native execution and timing workloads}
Quality outputs come from native execution with CPU fallback disabled. Runtime, backend and cached-reload checks verify the compiled context before evaluation. Model timing covers synchronous inference calls and excludes model loading, compilation, transfer, preprocessing, warm-up and display.

\emph{Source64} reports the median of three 64-source-image session p50 values; full-target evaluation pools model calls per dataset. VENTUNO Q's earlier campaigns run sequential conditions except for the alternating MSE pair. Its matched six-condition campaign rotates conditions across chunks of up to 48 targets per cached session, with three source warm-ups. IQ9075 set B uses one session per condition/dataset. A separate VENTUNO Q sustained study runs each condition for 600 seconds on cyclic source inputs. These are separate, non-isolated protocols (Appendix~\ref{app:timing_checks}).

\section{Results}
\label{sec:results}
\subsection{Quality recovery from PTQ to VRQ-PTQ}
All four MinMax/MSE--source-set pairs improve the 3-set AUROC Mean (Table~\ref{tab:multidata}): +0.902/+0.659\,pp in A and +0.738/+0.522\,pp in B. Relative to HTP FP16 (88.663), the set A gains restore 98\% of MinMax's 0.917\,pp loss and 90\% of MSE's 0.735\,pp loss. This within-observer recovery is the central result; the best ordinary observer varies by metric and dataset.
\begin{table*}[t]
\centering\footnotesize
\caption{Image and original-mask Pixel AUROC ($\times100$). Calibration sets A and B use their own matched source/export configurations. The final column weights Image/Pixel equally within each dataset and datasets equally. Bold marks the highest displayed aggregate and corresponding condition within each set; floating-point references are not ranked. Complete dataset means and the alternative KL implementation appear in Supplementary Table~\ref{tab:complete_auroc}.}
\label{tab:multidata}\label{tab:adaround}
\setlength{\tabcolsep}{3pt}
\begin{adjustbox}{max width=\linewidth}
\begin{tabular}{llrrrrrrr}
\toprule
 & & \multicolumn{2}{c}{\textbf{MVTec AD}} & \multicolumn{2}{c}{\textbf{VisA}} & \multicolumn{2}{c}{\textbf{AD2 public}} & \textbf{3-set}\\
\textbf{Condition} & \textbf{Set} & \textbf{Image} & \textbf{Pixel} & \textbf{Image} & \textbf{Pixel} & \textbf{Image} & \textbf{Pixel} & \textbf{AUROC Mean}\\
\midrule
GPU FP32 & A & 96.445 & 92.050 & 92.888 & 95.630 & 67.705 & 87.230 & 88.658 \\
HTP FP16 & A & 96.448 & 92.057 & 92.932 & 95.637 & 67.650 & 87.252 & 88.663 \\
\midrule
MinMax PTQ & A & 96.047 & 91.990 & 91.565 & 95.503 & 64.090 & 87.282 & 87.746 \\
MinMax PTQ & B & 96.236 & 91.973 & 91.482 & 95.477 & 64.725 & 87.360 & 87.876 \\
MinMax PTQ + VRQ (override) & A & 96.463 & 92.062 & 92.862 & 95.608 & 67.389 & 87.225 & 88.602 \\
\textbf{MinMax PTQ + VRQ} & A & 96.529 & 92.069 & 93.001 & 95.609 & 67.440 & 87.241 & \textbf{88.648} \\
\textbf{MinMax PTQ + VRQ} & B & 96.463 & 92.065 & 92.876 & 95.623 & 67.428 & 87.229 & \textbf{88.614} \\
\midrule
Entropy PTQ & A & 96.326 & 92.004 & 92.118 & 95.523 & 65.395 & 87.377 & 88.124 \\
Entropy PTQ + VRQ & A & 96.523 & 92.079 & 92.786 & 95.595 & 65.913 & 87.293 & 88.365 \\
\midrule
MSE PTQ & A & 96.232 & 91.994 & 91.553 & 95.501 & 64.999 & 87.289 & 87.928 \\
MSE PTQ & B & 96.378 & 91.995 & 91.687 & 95.513 & 65.280 & 87.374 & 88.038 \\
MSE PTQ + VRQ (override) & A & 96.412 & 92.064 & 93.024 & 95.607 & 67.434 & 87.208 & 88.625 \\
MSE PTQ + VRQ & A & 96.516 & 92.066 & 92.858 & 95.607 & 67.263 & 87.212 & 88.587 \\
MSE PTQ + VRQ & B & 96.422 & 92.065 & 92.832 & 95.617 & 67.171 & 87.249 & 88.560 \\
\midrule
AdaRound (fixed QNN grids) & A & 96.443 & 92.033 & 91.949 & 95.508 & 64.876 & 87.333 & 88.024 \\
 
\bottomrule
\end{tabular}
\end{adjustbox}
\end{table*}

\begin{figure}[!tbp]
\centering
\includegraphics[width=\linewidth]{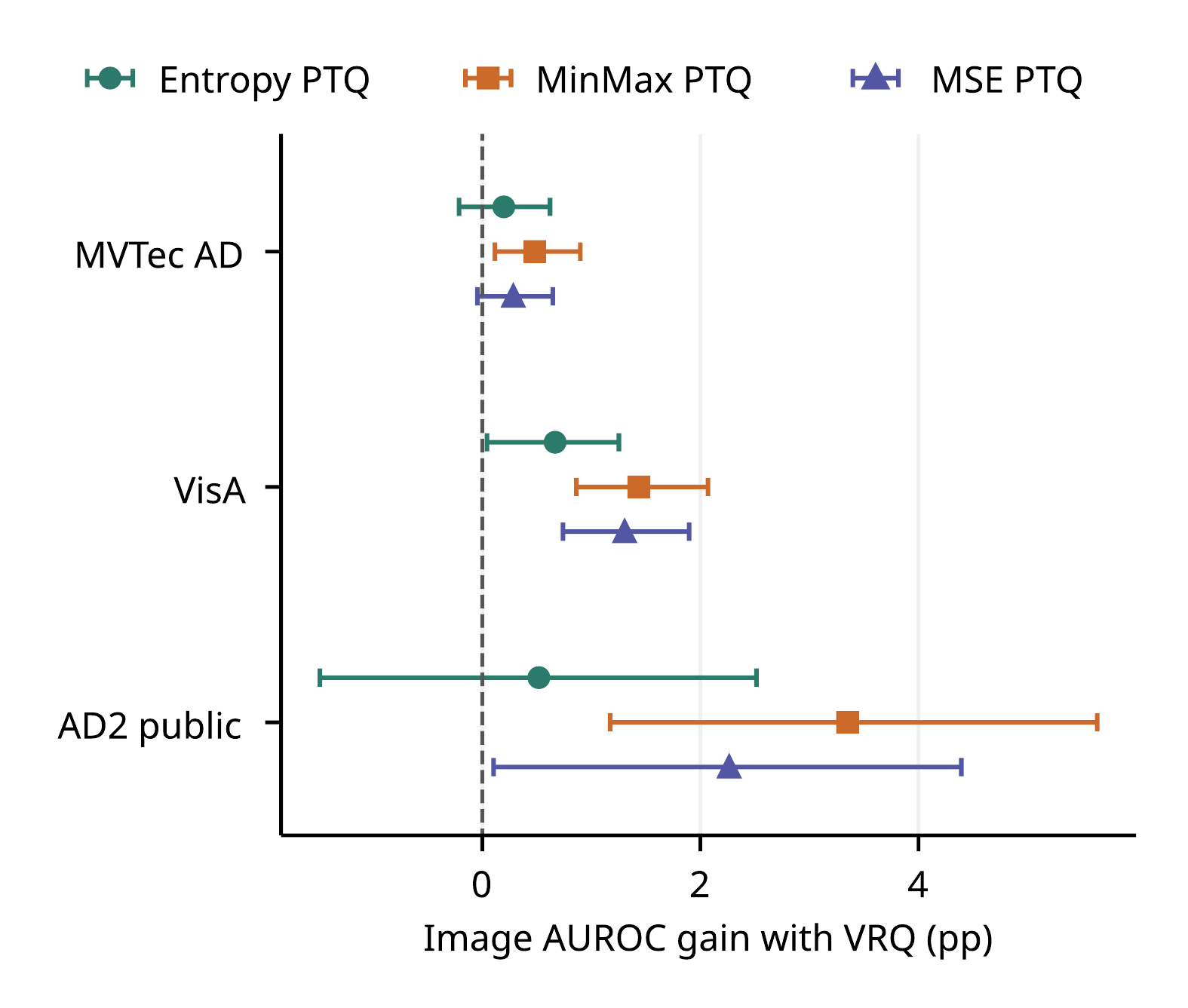}
\caption{Calibration set A paired macro Image AUROC gains from adding VRQ to Entropy, MinMax, and MSE PTQ, with 95\% percentile intervals and fixed categories. Each contrast uses its own matched baseline. Resampling uses images for MVTec AD/VisA and scene clusters for AD2. Zero denotes no change; intervals exclude calibration-seed variation.}
\label{fig:bootstrap}
\end{figure}

Set A's paired cluster-jackknife 95\% intervals are [+0.456, +1.348]\,pp for MinMax and [+0.244, +1.074]\,pp for MSE. Image-level discrimination carries the gain (+1.756/+1.285\,pp); Pixel AUROC changes by only +0.048/+0.033\,pp, with both intervals including zero (Supplementary Table~\ref{tab:cluster_recovery}). Baseline PTQ already lowers Pixel AUROC by at most 0.14\,pp per dataset, so localization recovery appears instead in region overlap. The separate dataset-level Image bootstrap analysis (Figure~\ref{fig:bootstrap}, Table~\ref{tab:bootstrap}) finds five of six MinMax/MSE intervals exclude zero; MSE on MVTec AD is the exception ([$-0.044$, +0.647]\,pp).

The 3-set AUPRO Mean gains are +1.280/+1.282\,pp for MinMax/MSE in A and +1.227/+1.218\,pp in B (Table~\ref{tab:aupro}). Both set A pairs recover 89\% of their loss relative to HTP FP16. This regional metric complements AUROC's score ordering by measuring defect coverage under constrained false positives.
\begin{table*}[t]
\centering\footnotesize
\caption{Original-mask AUPRO ($\times100$). The final column weights the 0.05/0.30 FPR limits equally within each dataset, then weights datasets equally. Bold retains the within-set maxima among quantized conditions; floating-point references are not ranked. This regional aggregate is separate from the AUROC mean. Table~\ref{tab:complete_aupro} lists additional set B controls.}
\label{tab:aupro}
\setlength{\tabcolsep}{3pt}
\begin{adjustbox}{max width=\linewidth}
\begin{tabular}{llrrrrrrr}
\toprule
 & & \multicolumn{2}{c}{\textbf{MVTec AD}} & \multicolumn{2}{c}{\textbf{VisA}} & \multicolumn{2}{c}{\textbf{AD2 public}} & \textbf{3-set}\\
\textbf{Condition} & \textbf{Set} & \textbf{@0.30} & \textbf{@0.05} & \textbf{@0.30} & \textbf{@0.05} & \textbf{@0.30} & \textbf{@0.05} & \textbf{AUPRO Mean}\\
\midrule
GPU FP32 & A & 88.649 & 72.099 & 89.598 & 66.187 & 50.651 & 25.790 & 65.496 \\
HTP FP16 & A & 88.641 & 72.071 & 89.620 & 66.232 & 50.656 & 25.792 & 65.502 \\
\midrule
MinMax PTQ & A & 88.152 & 70.819 & 88.946 & 63.667 & 48.898 & 23.938 & 64.070 \\
MinMax PTQ & B & 88.186 & 70.878 & 88.815 & 63.419 & 49.334 & 24.165 & 64.133 \\
MinMax PTQ + VRQ (override) & A & 88.589 & 71.940 & 89.468 & 65.888 & 50.470 & 25.683 & 65.340 \\
\textbf{MinMax PTQ + VRQ} & A & 88.595 & \textbf{71.948} & \textbf{89.477} & \textbf{65.919} & 50.481 & 25.681 & \textbf{65.350} \\
\textbf{MinMax PTQ + VRQ} & B & 88.582 & 71.929 & 89.504 & 65.899 & 50.507 & \textbf{25.738} & 65.360 \\
\midrule
Entropy PTQ & A & 88.164 & 70.700 & 88.901 & 63.464 & 49.008 & 23.891 & 64.021 \\
\textbf{Entropy PTQ + VRQ} & A & \textbf{88.609} & 71.945 & 89.471 & 65.864 & \textbf{50.565} & 25.633 & 65.348 \\
\midrule
MSE PTQ & A & 88.154 & 70.790 & 88.890 & 63.576 & 48.938 & 23.996 & 64.057 \\
MSE PTQ & B & 88.218 & 70.920 & 88.853 & 63.314 & 49.323 & 24.263 & 64.149 \\
MSE PTQ + VRQ (override) & A & 88.592 & 71.937 & 89.464 & 65.861 & 50.459 & 25.675 & 65.331 \\
\textbf{MSE PTQ + VRQ} & A & 88.588 & 71.942 & 89.470 & 65.885 & 50.461 & \textbf{25.687} & 65.339 \\
\textbf{MSE PTQ + VRQ} & B & \textbf{88.590} & \textbf{71.938} & \textbf{89.509} & \textbf{65.901} & \textbf{50.537} & 25.729 & \textbf{65.367} \\
\midrule
AdaRound (fixed grids) & A & 88.214 & 70.884 & 89.030 & 63.904 & 49.025 & 23.845 & 64.150 \\
 
\bottomrule
\end{tabular}
\end{adjustbox}
\end{table*}

MinMax PTQ+VRQ leads the observed AUROC aggregate in both sets. The AUPRO leader changes from MinMax in A to MSE in B, with small gaps (0.011/0.007\,pp) relative to the within-pair gains. Across six MVTec calibration repeats (64/128 source images, three seeds), MinMax Image gains remain positive (+0.126 to +0.592\,pp), and corrected models vary less than baselines (range 0.110 versus 0.419\,pp; Table~\ref{tab:repeat}).

Entropy+VRQ also improves Image AUROC and AUPRO@0.05 on all three datasets. Category failures remain: AD2 rice loses $-7.116$\,pp in Image AUROC despite improved Pixel AUROC. Appendix~\ref{app:secondary} reports category results and the separate Kullback--Leibler (KL) and Percentile controls. The KL implementation's set-B AUROC aggregate changes only from 74.068 to 74.758: Value correction leaves other baseline failures unresolved.

\subsection{Tensor overrides and added QDQ}
\label{sec:boundary_controls}\label{sec:independent_results}
The MinMax same-grid control adds QDQ while retaining the inherited Value range. Its outputs match baseline PTQ across all 15 MVTec categories. The benefit therefore depends on changing the range, rather than inserting nodes alone.

Table~\ref{tab:independent_v} compares Value assignments on MVTec AD. Under Entropy PTQ, independently observing Value with Entropy raises AUPRO@0.05 by +1.235\,pp, close to the +1.245\,pp gain from the source-MinMax added-QDQ correction. Assigning the matched MinMax grid through a tensor override gives the highest localization point estimates in this block. These controls reproduce recovery through independent range assignment using the standard quantizer interface.
\begin{table*}[t]
\centering\footnotesize
\caption{Independent Value controls on MVTec AD, original masks ($\times100$). A--D label the Entropy controls. B independently observes Value with Entropy; C and D share the source-MinMax Value grid. Within each observer block, bold Image/Pixel pairs identify the largest equal-weight AUROC mean; bold AUPRO cells identify column maxima. The MinMax/MSE block uses calibration set A on VENTUNO Q.}
\label{tab:independent_v}
\setlength{\tabcolsep}{3pt}
\begin{adjustbox}{max width=\linewidth}
\begin{tabular}{>{\raggedright\arraybackslash}p{34mm}>{\raggedright\arraybackslash}p{37mm}rrrr}
\toprule
\textbf{Condition} & \textbf{Value rule / realization} & \textbf{Image} & \textbf{Pixel} & \textbf{AUPRO@0.05} & \textbf{AUPRO@0.30}\\
\midrule
A: Entropy PTQ & Shared QKV grid & 96.3259 & 92.0039 & 70.6999 & 88.1637 \\
B: Entropy PTQ + independent V & Entropy / tensor override & 96.4045 & 92.0727 & 71.9350 & 88.6047 \\
C: Entropy PTQ + VRQ (override) & Source MinMax / tensor override & 96.4538 & 92.0793 & \textbf{71.9653} & \textbf{88.6165} \\
\textbf{D: Entropy PTQ + \vr{}} & Source MinMax / extra QDQ & \textbf{96.5234} & \textbf{92.0788} & 71.9452 & 88.6088 \\
 
\midrule
MinMax PTQ & Shared QKV grid & 96.0473 & 91.9900 & 70.8187 & 88.1515 \\
MinMax PTQ + VRQ (override) & Source MinMax / tensor override & 96.4626 & 92.0619 & 71.9399 & 88.5892 \\
\textbf{MinMax PTQ + VRQ} & Source MinMax / extra QDQ & \textbf{96.5286} & \textbf{92.0691} & \textbf{71.9476} & \textbf{88.5947} \\
\midrule
MSE PTQ & Shared QKV grid & 96.2317 & 91.9942 & 70.7903 & 88.1545 \\
MSE PTQ + VRQ (override) & Source MinMax / tensor override & 96.4124 & 92.0643 & 71.9370 & \textbf{88.5921} \\
\textbf{MSE PTQ + VRQ} & Source MinMax / extra QDQ & \textbf{96.5164} & \textbf{92.0661} & \textbf{71.9416} & 88.5879 \\
 
\bottomrule
\end{tabular}
\end{adjustbox}
\end{table*}

The three-dataset MinMax and MSE override controls reach the same conclusion. Their 3-set AUROC Mean gains over PTQ are +0.855\,pp [+0.404, +1.307] and +0.697\,pp [+0.283, +1.111], respectively. Added QDQ minus the matched tensor override is +0.047\,pp [$-0.098$, +0.192] for MinMax and $-0.038$\,pp [$-0.179$, +0.103] for MSE. Both intervals include zero.

\subsection{Comparison with fixed-grid weight rounding}

Set A AdaRound changes weight rounding while retaining Entropy PTQ's weight scales and activation grids; Entropy PTQ+VRQ is its matched comparison. VRQ exceeds this control in Image AUROC and AUPRO@0.05 on all datasets, although Image bootstrap intervals exclude zero only on VisA. Pixel AUROC is higher on MVTec/VisA and lower on AD2. The AUROC/AUPRO aggregates are 88.024/64.150 for AdaRound versus 88.365/65.348 for VRQ (Tables~\ref{tab:complete_auroc} and~\ref{tab:aupro}). This comparison concerns the specified fixed-grid implementation.

\subsection{Source-fixed operating points}
Each condition flags the same 30 of 608 source normal images (4.934\%), but target FNR/FPR changes (Table~\ref{tab:operating}). On MVTec and VisA, PTQ produces fewer alarms and more misses; VRQ reverses that shift. For example, MVTec MinMax FNR/FPR moves from 21.25/7.64\% to 19.96/11.21\%, versus 20.29/10.09\% for HTP FP16. On AD2, every condition misses about 80\% of anomalies; MinMax and MSE slightly reduce FNR but move FPR in opposite directions. Thus lower baseline FPR reflects an operating-point shift, not necessarily better discrimination. Entropy+VRQ yields 29 fewer false negatives and 10 additional false positives on MVTec, and 20 fewer/9 additional on VisA.
\begin{table*}[t]
\centering\small
\caption{Target operating points with one threshold per condition fixed on 608 source normal images. Entries are category-macro FNR/FPR (\%, lower is better), not accuracy or a receiver operating characteristic comparison at matched target FPR. All conditions use the same source pool and threshold rule, and each flags 30 of the 608 source normal images (empirical source FPR 4.934\%). Bold entries mark the lowest FNR or FPR per column among quantized conditions; FP references are not ranked. All non-FP32 results are native VENTUNO Q outputs. AD2 uses the public split; lighting views can share a scene.}
\label{tab:operating}
\setlength{\tabcolsep}{7pt}
\begin{adjustbox}{max width=\linewidth}
\begin{tabular}{lrrrrrr}
\toprule
 & \multicolumn{2}{c}{\textbf{MVTec AD}} & \multicolumn{2}{c}{\textbf{VisA}} & \multicolumn{2}{c}{\textbf{AD2 public}}\\
\textbf{Condition} & \textbf{FNR} & \textbf{FPR} & \textbf{FNR} & \textbf{FPR} & \textbf{FNR} & \textbf{FPR}\\
\midrule
GPU FP32 & 19.86 & 10.69 & 46.83 & 10.25 & 80.42 & 1.40 \\
HTP FP16 & 20.29 & 10.09 & 47.50 & 10.25 & 80.90 & 1.04 \\
MinMax PTQ & 21.25 & 7.64 & 47.92 & 9.67 & 80.08 & 2.02 \\
MinMax PTQ + VRQ (override) & 20.73 & 9.16 & 47.08 & 10.42 & 80.88 & \textbf{1.40} \\
MinMax PTQ + \vr{} & 19.96 & 11.21 & 46.42 & 10.42 & 79.64 & 1.82 \\
Entropy PTQ & 21.73 & 7.55 & 48.17 & 9.58 & 80.38 & 2.26 \\
Entropy PTQ + \vr{} & 19.92 & 10.04 & 46.50 & 10.33 & 80.20 & 1.96 \\
AdaRound (fixed grids) & 20.66 & 8.22 & 48.00 & 9.42 & 80.61 & 3.18 \\
MSE PTQ & 22.23 & \textbf{5.48} & 48.50 & \textbf{9.08} & 80.87 & 1.55 \\
MSE PTQ + VRQ (override) & 19.89 & 10.17 & 46.67 & 10.50 & 80.16 & 1.61 \\
MSE PTQ + VRQ & \textbf{19.44} & 10.70 & \textbf{45.58} & 10.75 & \textbf{79.38} & 2.17 \\
 
\bottomrule
\end{tabular}
\end{adjustbox}
\end{table*}

Decision preservation and net recovery distinguish two populations; per-image decision counts are reported for the initially evaluated Entropy pair. Among 986 MVTec defects detected by the floating-point reference, Entropy PTQ misses 38 and its VRQ variant misses 4. Over the complete anomalous set, VRQ recovers 36 detections and loses 7. The operating-point improvement therefore accompanies a false-alarm trade-off rather than removing every detector error. Supplementary examples show both recoveries and new misses under the unchanged source-threshold rule.

\subsection{Native execution cost and reproducibility}
Table~\ref{tab:devices} separates repeated-source and full-target timings. On IQ9075, MinMax PTQ and its VRQ variant have source64 model p50 values of 364.31 and 376.75\,ms; the floating-point reference takes 797.91\,ms. Under this protocol the corrected model retains a 2.12$\times$ speedup with 3.41\% overhead over baseline PTQ. In set B's MVTec full-target evaluation, the pair takes 603.66 and 641.03\,ms, a 6.19\% increment. Across the MinMax and MSE pairs on all three boards, the increment ranges from 2.57\% (RB5 Gen2, source64, MinMax) to 8.11\% (RB5 Gen2, AD2 full-target, MSE; Supplementary Table~\ref{tab:complete_devices}). These measurements characterize distinct workloads and session settings; their absolute times are not interchangeable.
\begin{table*}[t]
\centering\footnotesize
\caption{Model p50 (ms) under distinct workloads. Source64 reports the median of three 64-source-image session medians. Full-target values cover all targets in each dataset; set A uses the rotated-order six-condition evaluation and set B the single-session IQ9075 evaluation. Compile, transfer, preprocessing and warm-up are excluded. These workloads are not pooled. Bold timing cells retain the within-protocol minima. Complete device records are in Supplementary Table~\ref{tab:complete_devices}.}
\label{tab:devices}\label{tab:full_target_timing}
\begin{tabular}{lrrr}
\toprule
\multicolumn{4}{l}{\textbf{Repeated source64 workload}}\\
\textbf{Board} & \textbf{HTP FP16} & \textbf{MinMax PTQ} & \textbf{MinMax PTQ+VRQ}\\
\midrule
IQ9075 & 797.91 & \textbf{364.31} & 376.75\\
RB5 Gen2 & 1040.58 & \textbf{545.34} & 559.35\\
\bottomrule
\end{tabular}
\par\vspace{8pt}
\begin{adjustbox}{max width=\linewidth}
\begin{tabular}{lllrrr}
\toprule
\multicolumn{6}{l}{\textbf{Full-target model calls}}\\
\textbf{Board} & \textbf{Set} & \textbf{Condition} & \textbf{MVTec AD} & \textbf{VisA} & \textbf{AD2 public}\\
\midrule
VENTUNO Q & A & MinMax PTQ & 385.53 & 385.69 & 385.42 \\
VENTUNO Q & A & MinMax PTQ + VRQ (override) & 398.42 & 398.22 & 398.18 \\
VENTUNO Q & A & MinMax PTQ + VRQ & 398.16 & 398.29 & 398.47 \\
VENTUNO Q & A & MSE PTQ & \textbf{385.17} & \textbf{385.53} & \textbf{385.24} \\
VENTUNO Q & A & MSE PTQ + VRQ (override) & 398.00 & 398.49 & 398.27 \\
VENTUNO Q & A & MSE PTQ + VRQ & 398.54 & 398.18 & 397.97 \\
IQ9075 & B & MinMax PTQ & \textbf{603.66} & 603.68 & \textbf{604.18} \\
IQ9075 & B & MinMax PTQ + VRQ & 641.03 & 640.45 & 640.58 \\
IQ9075 & B & MSE PTQ & 604.36 & \textbf{603.51} & 604.45 \\
IQ9075 & B & MSE PTQ + VRQ & 641.79 & 640.68 & 641.21 \\
 
\bottomrule
\end{tabular}
\end{adjustbox}
\end{table*}

The six-condition VENTUNO Q sustained study completes 8,950 measured calls with QNN-only execution and bit-exact repeated outputs. MinMax PTQ+VRQ has model p50/p95 of 400.37/404.03\,ms, versus 385.27/388.90\,ms for baseline PTQ; MSE PTQ+VRQ has 397.17/400.86\,ms, versus 386.25/389.87\,ms. Table~\ref{tab:sustained_recovery} reports all conditions and within-run windows.

Override and added-QDQ latency orderings vary across observers and workloads (Tables~\ref{tab:devices}, \ref{tab:sustained_recovery}). On RB5 Gen2, same-grid QDQ takes 543.66\,ms versus 545.34\,ms for baseline PTQ, while override/added-QDQ corrections take 560.66/559.35\,ms (Table~\ref{tab:controls}). The increment accompanies the independent Value grid rather than node insertion; its origin within the compiled graph remains unresolved.

The combined context-binary and wrapper size decreases from 690,576,266 to 357,191,402 bytes for the evaluated quantized deployment, with unchanged parameter count. PTQ supplies the principal storage and speed benefits. The matched set-B MSE pair reproduces the displayed image, pixel and region metrics across IQ9075, VENTUNO Q and RB5 Gen2, while each board records its own latency. This establishes reproducibility within the shared runtime family. Comparisons with other detectors are retained as application context in Appendix~\ref{app:detector_context}; their backbone and training differences are separate from VRQ's within-detector effect.

\section{Discussion and Limitations}
\subsection{What VRQ changes}
The controls distinguish choosing a calibration objective from assigning its range to the consumed tensor. Q/K ranges match the first-block observations, whereas Value inherits a wider projection grid. Correcting Value improves native output; same-grid node insertion leaves it unchanged. Full-detector comparisons connect this intervention to discrimination and region-overlap recovery.

Tensor-override controls reproduce recovery without added QDQ, with similar latency increments and between-realization quality intervals that include zero. Independent source-calibrated range assignment at PV is the useful principle. Better upstream reconstruction alone need not recover anomaly ranking: tensor reconstruction, compiled fidelity and score ordering are distinct objectives.

\subsection{Engineering value and deployment choice}
For high-mix, low-volume inspection, source-only calibration reduces recurring model-preparation work when products change. Reusing a detector without target-specific retraining or quantization calibration addresses one part of the adoption burden. Camera setup, decision thresholds and acceptance criteria still require validation at the installation site. Equipment cost, power and commissioning effort motivate deployment; quantifying those benefits requires measurements beyond the quality, latency and storage evaluation reported here.

PTQ supplies the main speed and storage savings; VRQ restores 90--98\% of the AUROC loss and about 89\% of the AUPRO loss in set A, at approximately 3--8\% additional MinMax/MSE model latency. MinMax+VRQ leads the observed AUROC aggregate in both source sets, while regional rankings vary. Deployment choice should therefore combine discrimination, localization and operating-point requirements with native timing.

\subsection{Applicability across models and runtimes}
Validation uses the released AnomalyVFM/RADIO detector and the QNN/HTP runtime family, with weights fixed within each matched pair. The datasets test recovery across inspection targets, and the boards test reproducibility within this runtime family. Other-detector comparisons provide application context; cross-model and cross-architecture validation of VRQ remains open.

Transfer depends on the presence of the diagnosed range mismatch and the runtime's treatment of tensor encodings. For another model, exporter or backend, the procedure is to trace the Value encoding to PV, compare its range with source observations, and verify an independent range assignment under native execution. The mechanism predicts limited benefit when Value already receives an appropriate independent encoding; testing that prediction requires additional models and runtimes.

\subsection{Limitations}
\label{sec:limits}
The all-block audit establishes encoding propagation; Q/K numerical matching is checked in the first block and native probes cover selected depths. Full-model internal precision and W8A8 operation remain untested. The intervals establish no preferred implementation.

Intervals condition on fixed calibration inputs and categories, without multiplicity adjustment. Dataset Image intervals use paired bootstrap; Image/Pixel aggregates use cluster jackknife. Regional metrics, operating points and set B's aggregate lack corresponding intervals. AdaRound uses one optimization seed. Exploratory bottle/toothbrush observer and rotation rankings need not transfer to full datasets.

Source labels balance normal/anomalous calibration examples; zero-shot refers to unseen inspection targets. MinMax/MSE and independent-Value controls followed the initial Entropy comparison, making the multi-observer synthesis subsequent analysis. Development repeatedly inspected MVTec results and source-to-target export checks. Later MSE calibration passes a bounded numerical audit rather than bit-exact reproduction (Appendix~\ref{app:a_mse_reproduction}).

Timing is non-isolated, with background services stopped for matched evaluations. Source64, full-target and sustained workloads remain separate; their latency differences are not causally isolated. Each sustained condition has one run, and target timing lacks between-session variability estimates. Original-mask and native-grid pixel metrics remain separate. Category and lighting failures persist; clipping and source-to-target shifts warrant further study~\cite{eaai_bipolar}.

\section{Conclusion}
VRQ corrects inherited Value ranges in the deployed AnomalyVFM/RADIO detector while preserving PTQ weights and ordinary activation encodings. Matched MinMax/MSE comparisons improve three-dataset discrimination and region overlap, recovering 90--98\% of the primary calibration set's AUROC loss at approximately 3--8\% additional model latency. Tensor-override and native controls locate the benefit in independent range assignment rather than node insertion or declared bit-width. The result is a diagnosed, implementable correction of deployment loss, with quality and cost verified across the tested Qualcomm runtime family.

\section*{Data availability}
The study uses the public MVTec AD, VisA, MVTec AD 2 public split and Real-IAD datasets~\cite{mvtec,visa,mvtecad2,realiad}. Dataset access is subject to the respective providers' terms. Code and supporting reproducibility materials will be released.

\section*{Acknowledgments}
This work was supported by the Technology Innovation Program(Semiconductor Advanced Industry Technology Development Project)(RS-2026-25522959, Development of Multimodal On-sensor PIM-AI Semiconductor and Sensory Data-based Intelligent Motion Control Technology for Humanoid Robot Hands) funded By the Ministry of Trade Industry \& Energy(MOTIE, Korea). The authors acknowledge support from Qualcomm Technologies, Inc. through the Qualcomm AI Program for Innovators 2026--APAC. Plaid Labs Inc. provided access to GPU servers used for the experiments reported in this study.

\FloatBarrier

\par\medskip

\useRomanappendicesfalse
\appendices
\setcounter{figure}{0}
\renewcommand{\thefigure}{S\arabic{figure}}
\renewcommand{\theHfigure}{S\arabic{figure}}
\begin{table}[!tbp]\centering\textbf{Nomenclature}\par\medskip
\begin{center}\small
\begin{tabularx}{\linewidth}{lX}
\toprule
\textbf{Term} & \textbf{Meaning}\\
\midrule
AUPRO & Area under the per-region overlap curve\\
AUROC & Area under the receiver operating characteristic curve\\
CPU / GPU & Central / graphics processing unit\\
FP / FP16 / FP32 & Floating point / 16-bit / 32-bit floating point\\
FNR / FPR & False-negative / false-positive rate\\
HTP & Hexagon Tensor Processor\\
KL & Kullback--Leibler divergence\\
MSE & Mean squared error\\
NPU & Neural processing unit\\
ONNX & Open Neural Network Exchange\\
PTQ & Post-training quantization\\
PV & Probability--value multiplication\\
QDQ & Quantize--dequantize pair\\
QKV & Query--key--value\\
QNN & Qualcomm Neural Network execution stack\\
RADIO & Reduce All Domains Into One\\
RMS / RSS & Root mean square / resident set size\\
UINT8 / UINT16 & Unsigned 8-bit / 16-bit integer encoding\\
ULP & Unit in the last place\\
ViT & Vision transformer\\
VRQ & Value-Range Quantization\\
W8A16 / W8A8 & 8-bit weights with 16-bit / 8-bit activation encoding\\
\bottomrule
\end{tabularx}
\end{center}\end{table}

\renewcommand{\thesection}{\Alph{section}}

\section{Additional Evaluation Results}
\label{app:additional}
\setcounter{table}{0}
\renewcommand{\thetable}{A\arabic{table}}
\renewcommand{\theHtable}{A\arabic{table}}

\subsection{Paired intervals and detection decisions}
Tables~\ref{tab:bootstrap} and~\ref{tab:retention} detail uncertainty and defect decisions summarized in the main text.

\begin{table*}[t]
\centering\small
\caption{Paired \vr{}--PTQ differences (pp): Image 95\% percentile intervals from 2,000 paired resamples (AD2 scene clusters); Pixel point estimates. Intervals are descriptive and multiplicity-unadjusted.}
\label{tab:bootstrap}
\setlength{\tabcolsep}{5pt}
\begin{adjustbox}{max width=\linewidth}
\begin{tabular}{llrrrr}
\toprule
\textbf{Dataset} & \textbf{Observer} & \textbf{$\Delta$Image} & \textbf{Image 95\% interval} & \textbf{$\Delta$Pixel} & \textbf{Improved/tied/worse}\\
\midrule
MVTec AD & Entropy & +0.198 & [$-0.213$, +0.622] & +0.075 & 6/2/7 \\
MVTec AD & MinMax & +0.481 & [+0.116, +0.899] & +0.079 & 9/2/4 \\
MVTec AD & MSE & +0.285 & [$-0.044$, +0.647] & +0.072 & 7/4/4 \\
VisA & Entropy & +0.668 & [+0.044, +1.253] & +0.072 & 10/0/2 \\
VisA & MinMax & +1.436 & [+0.863, +2.070] & +0.106 & 9/0/3 \\
VisA & MSE & +1.306 & [+0.740, +1.897] & +0.106 & 10/0/2 \\
AD2 public & Entropy & +0.519 & [$-1.488$, +2.515] & $-0.084$ & 5/0/3 \\
AD2 public & MinMax & +3.350 & [+1.172, +5.639] & $-0.041$ & 7/0/1 \\
AD2 public & MSE & +2.264 & [+0.104, +4.393] & $-0.077$ & 6/0/2 \\
 
\bottomrule
\end{tabular}
\end{adjustbox}
\end{table*}

\begin{table*}[t]
\centering\small
\caption{Source-fixed defect decisions: FP32-detected defects (left) and all target defects (right). E: Entropy; E+VRQ: Entropy PTQ+\vr{}. Counts are images; AD2 views share scenes. False alarms: Table~\ref{tab:operating}.}
\label{tab:retention}
\setlength{\tabcolsep}{5pt}
\begin{adjustbox}{max width=\linewidth}
\begin{tabular}{lrrrrr}
\toprule
 & \multicolumn{3}{c}{\textbf{Among FP32-detected defects}} & \multicolumn{2}{c}{\textbf{All target defects (E $\to$ E+VRQ)}}\\
\cmidrule(lr){2-4}\cmidrule(lr){5-6}
\textbf{Dataset} & \textbf{Total} & \textbf{E misses} & \textbf{E+VRQ misses} & \textbf{Recovered} & \textbf{Lost}\\
\midrule
MVTec AD & 986 & 38 & 4 & 36 & 7 \\
VisA & 638 & 38 & 9 & 37 & 17 \\
AD2 public & 144 & 4 & 2 & 8 & 6 \\
 
\bottomrule
\end{tabular}
\end{adjustbox}
\end{table*}

\subsection{Context from other detectors}
\label{app:detector_context}
LiZAD~\cite{lizad} and AnomalyCLIP~\cite{anomalyclip} use official code and public checkpoints on the same 4,971 targets: VisA-trained checkpoints for MVTec AD and MVTec-trained checkpoints for VisA/AD2 public, without target-based selection. LiZAD combines DINOv3, ViT-S/16, and MobileCLIP2 S0; AnomalyCLIP uses CLIP's ViT-L/14. Both retain official $518\times518$ inputs and postprocessing versus RADIO's $768\times768$, resizing only prediction maps to original-mask dimensions.

LiZAD's map-maximum Image score is our derived metric, not one reported by its authors; shared GPU workloads preclude latency ranking.

On AD2 public, the deployment pipeline trails AnomalyCLIP in AUPRO@0.05 (25.633 versus 26.595) and LiZAD in AUPRO@0.30 (50.565 versus 50.870); these multiplicity-unadjusted system comparisons support no state-of-the-art claim.

\begin{table*}[t]
\centering\small
\caption{Matched-target detector comparison: macro AUROC (upper), AUPRO (lower), $\times100$; original masks, AD2 public split. Training data, backbones, and resolutions differ; AnomalyVFM/RADIO rows share one checkpoint. $\dagger$ LiZAD uses our map-maximum Image score.}
\label{tab:detectors}
\setlength{\tabcolsep}{3pt}
\begin{adjustbox}{max width=\linewidth}
\begin{tabular}{llrrrrrr}
\toprule
 &  & \multicolumn{2}{c}{\textbf{MVTec AD}} & \multicolumn{2}{c}{\textbf{VisA}} & \multicolumn{2}{c}{\textbf{AD2 public}}\\
\textbf{Model} & \textbf{Execution} & \textbf{Image} & \textbf{Pixel} & \textbf{Image} & \textbf{Pixel} & \textbf{Image} & \textbf{Pixel}\\
\midrule
LiZAD$^{\dagger}$ & GPU & 84.402 & 86.309 & 77.840 & 94.484 & 59.162 & 83.411 \\
AnomalyCLIP & GPU & 91.553 & 91.077 & 81.971 & 95.475 & 52.194 & 82.467 \\
AnomalyVFM/RADIO & GPU FP32 & 96.445 & 92.050 & 92.888 & 95.630 & 67.705 & 87.230 \\
AnomalyVFM/RADIO (Entropy PTQ+VRQ) & QNN W8A16 & 96.523 & 92.079 & 92.786 & 95.595 & 65.913 & 87.293 \\
 
\midrule
Model & Execution & @0.30 & @0.05 & @0.30 & @0.05 & @0.30 & @0.05\\
\midrule
LiZAD$^{\dagger}$ & GPU & 82.996 & 65.707 & 84.729 & 56.056 & 50.870 & 25.456 \\
AnomalyCLIP & GPU & 85.298 & 63.390 & 86.571 & 62.333 & 49.849 & 26.595 \\
AnomalyVFM/RADIO & GPU FP32 & 88.649 & 72.099 & 89.598 & 66.187 & 50.651 & 25.790 \\
AnomalyVFM/RADIO (Entropy PTQ+VRQ) & QNN W8A16 & 88.609 & 71.945 & 89.471 & 65.864 & 50.565 & 25.633 \\
 
\bottomrule
\end{tabular}
\end{adjustbox}
\end{table*}

\subsection{Lighting-dependent failure}
\label{app:lighting}
Table~\ref{tab:rice} decomposes the AD2 public rice failure discussed in the main text into six lighting conditions.

\begin{table}[!tbp]
\centering\scriptsize
\caption{AD2 public rice Image AUROC by lighting: 7 normal $\times$ 15 anomalous images (105 pairs) per row. E: Entropy; E+VRQ: Entropy PTQ+\vr{}; $\Delta$ is in pp. The pooled category metric additionally includes cross-lighting pairs.}
\label{tab:rice}
\setlength{\tabcolsep}{3pt}
\begin{adjustbox}{max width=\linewidth}
\begin{tabular}{lrrrr}
\toprule
\textbf{Lighting} & \textbf{FP32} & \textbf{E} & \textbf{E+VRQ} & \textbf{$\Delta$}\\
\midrule
regular & 87.62 & 73.81 & 73.33 & $-0.48$ \\
overexposed & 88.57 & 76.67 & 63.33 & $-13.33$ \\
underexposed & 91.43 & 80.95 & 73.33 & $-7.62$ \\
shift\_1 & 91.43 & 86.67 & 76.67 & $-10.00$ \\
shift\_2 & 90.48 & 87.62 & 80.00 & $-7.62$ \\
shift\_3 & 89.52 & 87.62 & 76.67 & $-10.95$ \\
 
\bottomrule
\end{tabular}
\end{adjustbox}
\end{table}

With ties worth 0.5 credit, improved/worsened pair credit totals 204.5/473.5 over $42\times90=3{,}780$ pairs, yielding $-7.116$\,pp. This pooled change includes cross-lighting pairs; averaging Table~\ref{tab:rice}'s 105-pair within-lighting rows yields a different statistic.

\subsection{Calibration repeats and observer controls}
Tables~\ref{tab:repeat} and~\ref{tab:controls} use native96 Pixel AUROC, unlike the primary original-mask localization results.

\begin{table}[!tbp]
\centering\scriptsize
\caption{VENTUNO Q MinMax calibration repeats. Seed suffixes omit 2026. $\Delta I,\Delta P$: \vr{}--PTQ differences in Image and native96 Pixel AUROC (pp).}
\label{tab:repeat}
\setlength{\tabcolsep}{3pt}
\begin{adjustbox}{max width=\linewidth}
\begin{tabular}{rrrrrr}
\toprule
\textbf{$N$} & \textbf{Seed} & \textbf{PTQ Image} & \textbf{\vr{} Image} & \textbf{$\Delta I$} & \textbf{$\Delta P$}\\
\midrule
64 & 1002 & 96.192 & 96.510 & +0.318 & +0.093 \\
64 & 1003 & 96.283 & 96.475 & +0.192 & +0.088 \\
64 & 1004 & 95.920 & 96.512 & +0.592 & +0.104 \\
128 & 1002 & 96.339 & 96.464 & +0.126 & +0.089 \\
128 & 1003 & 96.287 & 96.480 & +0.193 & +0.082 \\
128 & 1004 & 96.155 & 96.402 & +0.247 & +0.106 \\
 
\bottomrule
\end{tabular}
\end{adjustbox}
\end{table}

\begin{figure}[!tbp]
\centering
\includegraphics[width=\linewidth]{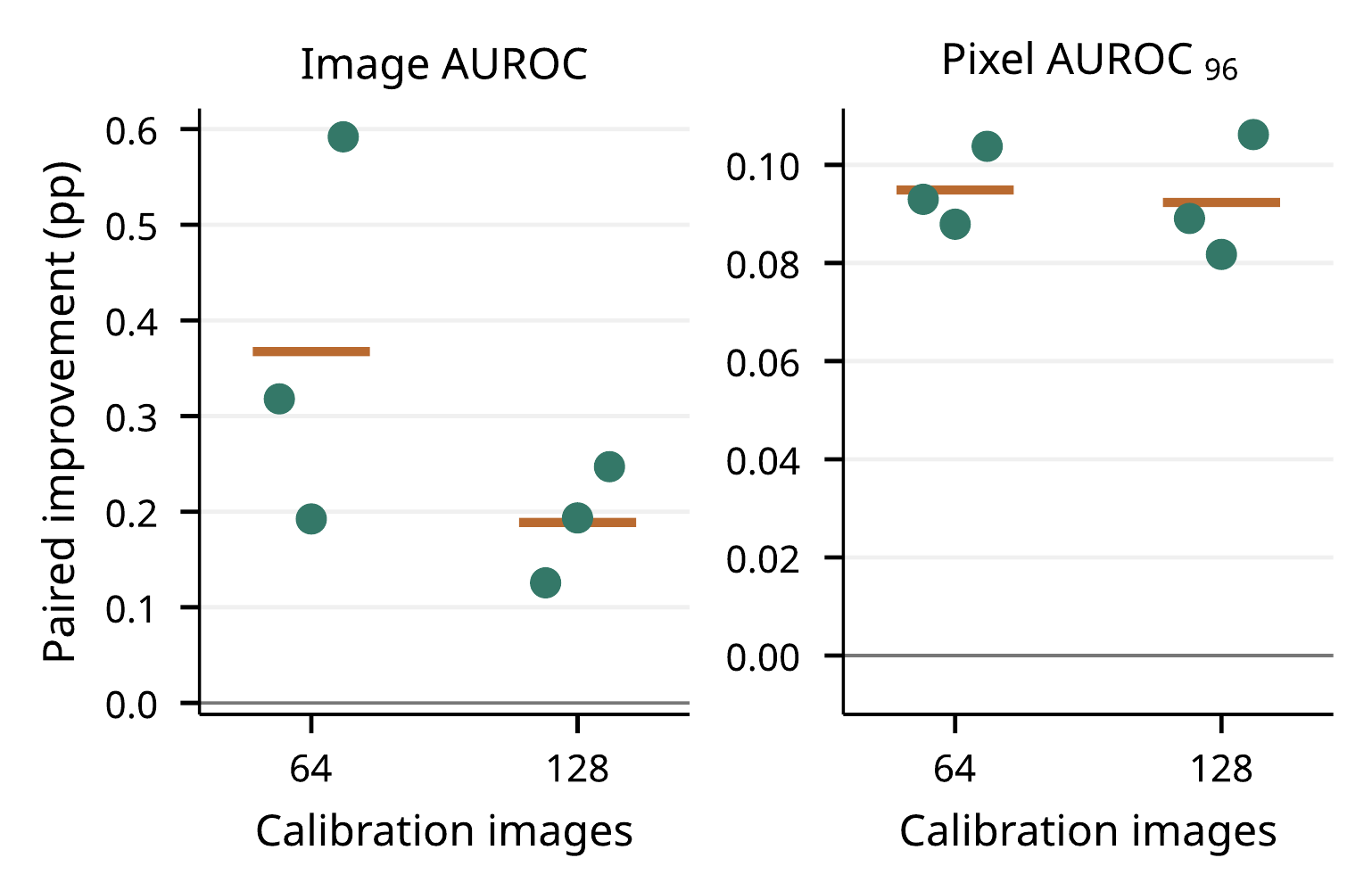}
\caption{Paired macro gains over MinMax PTQ across calibration repeats on the same MVTec AD test list. Points denote seeds; orange lines show three-seed means. Panels use different y-axis ranges.}
\label{fig:repeat}
\end{figure}

\paragraph{Calibration size and seeds}
Six MinMax PTQ/\vr{} pairs (64/128 calibration images, three seeds) are evaluated on the same 1,725 targets. All improve macro Image AUROC and \nativepixel{} (Table~\ref{tab:repeat}, Figure~\ref{fig:repeat}). For 64/128 images, Image gains are $+0.367\pm0.204$/$+0.189\pm0.061$\,pp and pixel gains are $+0.095\pm0.008$/$+0.092\pm0.013$\,pp (sample standard deviation across three seeds). Subsets can overlap; 128 images are not uniformly better. Individual conditions reduce Image AUROC in two to six categories, with a worst change of $-1.667$\,pp.

\begin{table}[!tbp]
\centering\small
\caption{RB5 Gen2 controls on 1,725 targets: macro AUROC and median of three source64 session p50 values.}
\label{tab:controls}
\setlength{\tabcolsep}{3pt}
\begin{adjustbox}{max width=\linewidth}
\begin{tabular}{lrrr}
\toprule
\textbf{Condition} & \textbf{Image} & \textbf{\shortstack{Pixel\\AUROC$_{96}$}} & \textbf{p50 ms}\\
\midrule
HTP FP16 & 96.455 & 92.058 & 1040.58 \\
MinMax PTQ & 96.047 & 91.992 & 545.34 \\
MinMax PTQ + \vr{} & 96.529 & 92.070 & 559.35 \\
Entropy PTQ & 96.326 & 92.006 & 548.28 \\
Entropy PTQ + \vr{} & 96.523 & 92.080 & 561.96 \\
Percentile PTQ & 72.437 & 91.356 & 553.95 \\
Percentile + \vr{} & 73.805 & 91.255 & 565.68 \\
MinMax same-grid & 96.047 & 91.992 & 543.66 \\
MinMax PTQ + VRQ (override) & 96.463 & 92.063 & 560.66 \\
 
\bottomrule
\end{tabular}
\end{adjustbox}
\end{table}

\section{Detailed Evaluation Protocols}
\label{app:protocols}
\subsection{Independent Value calibration checks}
\label{app:independent_protocol}
Rebuilt A matches baseline bytes; B/C preserve all 1,986 initializers and change only 48 Value quantize/dequantize scale/zero-point references; C/D grids match at all 24 boundaries. Each board passes four-condition source128 validation without CPU fallback and three bit-exact A/D anchor replays; B/C evaluate all targets, A/D reuse outputs.

The follow-up bootstrap uses label-stratified fixed categories and seed 20261019 without multiplicity adjustment; Section~\ref{sec:independent_results} interprets the intervals, excluding operating thresholds and isolated latency comparisons.

\paragraph{Matched MinMax/MSE extension}
The six-condition set A extension keeps source/target manifests, per-image input hashes, native worker, and original-mask evaluator. The four earlier control graphs reproduce archived Image/Pixel logits bit-for-bit on all 4,971 targets; both tensor-override graphs pass baseline-replay and initializer-preservation audits. Each condition compiles its own context, passes a three-image bit-exact cached-reload check without CPU fallback, and selects its threshold from the 608 source-normal images by the frozen rule. Chunks of at most 48 images use a fresh cached session and three warm-ups; order rotates by chunk and reverses on alternate six-chunk blocks. The competing NPU application is stopped during measurement and restored afterward; timing remains non-isolated, and campaigns are not averaged.

\subsection{Weight-rounding optimization and export}
\label{app:adaround_protocol}
In archived set A, AdaRound~\cite{adaround} retains Entropy PTQ's unsigned 8-bit integer (UINT8) weight scales, zero points, and axes. The Artificial Intelligence Model Efficiency Toolkit (AIMET) 2.31.0 optimizes 103 affine layers (6 convolution, 97 general matrix multiplication), 10,000 iterations each, with seed 20261006; disabled non-weight quantizers distinguish weight rounding from native-activation quantization-aware training.

Exported weights equal floor or floor+1 of their FP values on fixed grids; restoring all 103 tensors reproduces baseline bytes. Archived set A GPU-host optimization/export took 5,090.64 seconds (84.8 minutes), excluding setup, verification, and native evaluation; \vr{} instead requires observation and compilation.

Pair counting cross-checks Image AUROC, original-mask metrics cover all 35 categories, and AdaRound shares the source-only threshold rule.

\paragraph{Calibration set B implementation and audits}
Set B Entropy uses a frozen dense 2,048-bin histogram with a 128-bin KL search; all 461 activation thresholds reproduce the official ONNX Runtime calculation in a source-only check, but equivalence to set A Entropy is not established, so its lower scores describe this configuration only. AdaRound keeps the same AIMET version, seed, and 10,000 iterations per layer, fixes Entropy scales, zero points, axes, biases, activation grids, and topology, changing only UINT8 weight rounding; exports pass fixed-grid and floor-or-floor-plus-one audits. CPU-resident caches and ancestor-only capture graphs keep the relative RMS difference from the full FP graph below $1.30\times10^{-7}$ on two source images (tolerance, not bit-exact, parity).

\subsection{Original-mask region evaluation}
\label{app:region_protocol}
The per-region overlap-versus-FPR curve preserves distinct float32 scores, groups ties, avoids binning/pixel sampling, interpolates at the cutoff, and normalizes by the integration limit. Unit tests and MVTec evaluator 1.0~\cite{mvteceval} checks on two normal/two anomalous images per category--condition give maximum absolute difference $2.49\times10^{-9}$, reflecting official float32 FPR versus our float64 accumulation.

\subsection{Paired uncertainty and scene clustering}
\label{app:uncertainty_protocol}
Resampling shares indices across methods. AD2 fixes object positions across lighting~\cite{mvtecad2}; we group filename scene identifiers by category and label, verify lighting-set consistency, and do not treat 1,084 images as independent scenes.

Seeds are 20261006 (primary) and 20261013 (AdaRound); independent cumulative-rank checks reproduce intervals, and lighting/pair-order diagnostics select no ranges or thresholds.

The added set A MSE contrast uses saved native Image scores from the matched VENTUNO Q campaign, verified against source/target locks, without additional inference. It uses 2,000 paired label-stratified resamples, seed 20261006, fixed equal-weight categories, and the same image/scene units, keeping all lighting views together within each of the 184 AD2 scenes. Independent weighted receiver operating characteristic calculations cross-check 105 category--resample cases; the six archived Entropy/MinMax entries are retained at published precision. MSE draws, input identities, and random-number-generator ordering are in supplementary evidence; calibration variation and multiplicity correction are not included.

\subsection{Source-fixed threshold selection}
\label{app:threshold_protocol}
Each condition freezes its 578th sorted normal score, the $\lceil0.95\times608\rceil$th order statistic, treating only strictly greater scores as anomalous; the procedure, not numerical threshold, is shared, and all six original conditions plus AdaRound flag 30 source normal images.

Frozen thresholds determine true/false positives and negatives from target outputs; AdaRound uses the same pool/rule after context-hash and source-output checks. Target manifests only exclude duplicates; rates weight categories equally alongside totals.

\section{Deployment and Source Data}
\label{app:deployment}
\setcounter{table}{0}
\renewcommand{\thetable}{C\arabic{table}}
\renewcommand{\theHtable}{C\arabic{table}}
We use \href{https://huggingface.co/MaticFuc/anomalyvfm_radio/tree/17654e763c8fae5ae1c44e2ec421a427783d6196}{\nolinkurl{MaticFuc/anomalyvfm_radio}}, a large ViT with 16-pixel patches, loaded with the \href{https://github.com/MaticFuc/AnomalyVFM/tree/4da96b493a0b12e4e380fa5bf6573c126342855c}{pinned upstream code}.

\subsection{Native verification, timing, and preparation}
\label{app:timing_checks}
Only completed jobs enter aggregates after raw-output, input-count, model/context-hash, execution-provider, and completion-guard checks; separate code recomputes metrics.

Table~\ref{tab:runtime_schedule} separates archived VENTUNO Q target timings, source64 timings on other boards, and the set B full-target schedule. Source64 uses the median of three session p50 values on 64 source images; set B uses one cached-context session per method/dataset. These measurements characterize non-isolated device conditions.

The pilot alternates Entropy (A) and Entropy PTQ+\vr{} (B) in AB/BA/BA/AB order and stops/restores the competing NPU application; repeated inputs, residual CPU load, and unfixed power/temperature limit causal attribution.

\begin{table*}[!ht]
\centering\small
\caption{Latency schedules per condition; batch size one, warm-up excluded.}
\label{tab:runtime_schedule}
\setlength{\tabcolsep}{3pt}
\begin{adjustbox}{max width=\linewidth}
\begin{tabular}{L{.30\linewidth}L{.20\linewidth}L{.40\linewidth}}
\toprule
\textbf{Device / run} & \textbf{Warm-up} & \textbf{Timed calls}\\
\midrule
VENTUNO Q, primary & 3 per chunk session & 4,971 targets; pooled p50/p95\\
VENTUNO Q, pilot & 12 per session & $4\times64$; repeated source inputs\\
RB5 Gen2 / IQ9075, archived & 10 per session & $3\times64$ source images\\
Orin NX & 10 per session & $3\times64$ source images\\
\midrule
QNN boards, calibration set B & Sentinel check + 2 extra calls & 1,725 / 2,162 / 1,084 targets; one session per dataset\\
\bottomrule
\end{tabular}
\end{adjustbox}

\end{table*}
\begin{table}[!tbp]
\centering\small
\caption{VENTUNO Q pilot: model-only session p50 median and range (ms), 4 sessions per condition, repeating 3 source images. Ranges are not confidence intervals.}
\label{tab:latency_matched}
\setlength{\tabcolsep}{4pt}
\begin{adjustbox}{max width=\linewidth}
\begin{tabular}{lrr}
\toprule
\textbf{Condition} & \textbf{Median p50} & \textbf{Session p50 range}\\
\midrule
Entropy PTQ & 386.54 & 384.43--389.94 \\
Entropy PTQ + \vr{} & 398.27 & 397.50--398.77 \\
 
\bottomrule
\end{tabular}
\end{adjustbox}
\end{table}

Archived Qualcomm runs use ONNX Runtime 1.30.0 and QNN execution-provider (EP) plugin 2.6.0 (Python package versions, not SDK versions), HTP FP16 or W8A16, burst mode, profiling, disabled CPU fallback, and 2/1 intra-/inter-op threads; RB5 uses a host-side discovery shim. Orin NX uses TensorRT 10.3.0 with mixed FP16/FP32, disabled TensorFloat-32 (TF32), and MAXN power. QNN timing measures \texttt{session.run}; TensorRT timing includes synchronous host--device transfer. The eight pilot sessions (Table~\ref{tab:latency_matched}) are QNN-only with bit-exact repeated outputs.

The earlier 1,725-image MVTec-only MinMax run raises model p50 from \VentBaseTime{}\,ms to \VentVTime{}\,ms (\VentOverhead{}\% overhead), retaining \VentSpeedup{}$\times$ speedup over HTP FP16. For frozen MinMax PTQ+\vr{}, 256-image observation took 490.44\,s on eight CPU threads, and setup/compilation took 229.77\,s.

\subsection{Source partitions and selection}
Set A uses archived Real-IAD camera-1 (C1) views from 19 categories, with normal (OK) and anomalous (NG) directories. Verification of all 384 calibration/reserve files and 608 threshold-source files finds stored dimensions of $512\times512$; original byte and normalized-input hashes are preserved. Seed 20261001 selects distinct product groups for calibration (256 images, 128/128) and reserves (128, 64/64); probes use the first 16 reserves, and only calibration images fit ranges. Inputs use bilinear $768\times768$ resizing, channel--height--width layout, and float32/255 without augmentation.

Seed 20261006 selects 608 additional normal groups ($19\times32$), one C1 image each, excluding calibration, reserves, and repeats (Table~\ref{tab:repeat}). Calibration--target checks use file/decoded-RGB hashes; threshold selection additionally checks preprocessed-input hashes; target manifests only exclude duplicates.

\section{Calibration set A MSE protocol and numerical reproduction}\label{app:a_mse_reproduction}

The restored calibration passes a bounded FP numerical-drift audit rather than bit-exact reproduction.\footnote{The restored MinMax graph retains the topology, all 103 affine weight encodings and activation zero-points; activation scales differ by at most $1.15\times10^{-6}$ relatively (12 ULP). Published control models and table entries are unchanged.} Official re-encoding of the same FP biases accounts for 73 changed bias-scale vectors and 71 changed INT32 arrays. All 154 bias tensors re-encode exactly on both grids; decoded changes stay within half the summed old/new quantization steps plus roundoff. The maximum bias change is $3.05\times10^{-5}$ and maximum error to FP is $4.73\times10^{-5}$ on either grid. On two source inputs, all 812 observer tensors match the current official collector. Four held-out replays of the unchanged FP export satisfy the original parity tolerances despite small differences from archived outputs. The CPU/library cause remains unidentified.

The MSE pair alternates on each fixed input chunk, preserving input order and using separate fresh cached sessions with three identical source warmups. Timed model calls exclude transfer, compilation and warmup. In a separate 48-image MinMax diagnostic, stopping the concurrent NPU application reduces p50 from 787.13 to 388.27 ms with bit-identical logits. This diagnostic is distinct from full-target timings and does not establish isolated hardware performance.

\section{IQ9075 Observer and Rotation Ablations}
\label{app:bottle_followup}
\setcounter{table}{0}
\renewcommand{\thetable}{\Alph{section}\arabic{table}}
\renewcommand{\theHtable}{\Alph{section}\arabic{table}}

\subsection{Scope, metrics, and native timing}
This exploratory study uses the released checkpoint and code (Appendix~\ref{app:deployment}). Calibration uses 256 VisA images (12 categories, 128 normal and 128 anomalous), without label information. Campaigns cover the 83 MVTec AD bottle images (20/63) and later test frozen contexts on 42 toothbrush images. The later ablations were chosen after inspecting bottle results; these runs use a different source calibration from the 15-category evaluation.

Inputs are RGB, bilinearly resized to $768\times768$, float32/255, batch one; Image AUROC uses the raw image logit. \emph{Pixel96-NN} is pooled pixel AUROC over all 83 maps with nearest-neighbor $96\times96$ ground truth, distinct from \nativepixel{} and original-mask metrics. The FP reference is Image/Pixel96-NN 98.4127/93.5644.

IQ9075 uses ONNX Runtime 1.30.0, QNN EP 2.6.0 and HTP, with CPU fallback and graph-I/O offload disabled. Each fresh process verifies the cached context against compile-time output, warms up twice and times 83 synchronous calls, excluding loading, preprocessing and capture. Except in the initial weight control, p50 is the median of two session medians in mirrored order, with bit-identical outputs. These non-isolated campaigns differ from source64 and cannot resolve sub-millisecond effects.

\paragraph{Original-mask re-evaluation and toothbrush}
Bottle metrics are also recomputed with Equations~\eqref{eq:score}--\eqref{eq:map} and the original-mask protocol of Appendix~\ref{app:region_protocol} (67,230,000 pixels, 68 regions); second-implementation and MVTec-evaluator spot checks differ by less than $1.8\times10^{-11}$ in AUPRO. The six toothbrush contexts pass bit-exact reload checks, share no source-file or decoded-RGB hash with calibration images, and differ from the official evaluator by less than $2.5\times10^{-11}$. Toothbrush model p50 is 606.68--607.21\,ms without and 643.71--644.77\,ms with the added V boundary. VRQ changes toothbrush Image AUROC by three net pairs for MinMax and minus two for MSE (one pair is 0.2778\,pp; paired 95\% intervals $[-0.8333,3.6111]$ and $[-3.3333,1.6667]$\,pp). These VisA-calibrated graphs are not evaluated on full VisA because 142 calibration paths occur in its test split.

\subsection{Changing the calibration source}
Calibration set B is a fixed C1 selection from the original-resolution Real-IAD release, distinct from archived $512\times512$ set A: 256 calibration images (128/128, 19 categories) and 128 reserves, all from distinct product groups. File and decoded-RGB hashes find zero overlap with all 4,971 public targets, so the toothbrush evaluation is cross-dataset despite Real-IAD's toothbrush category. The fixed export is recalibrated into six MinMax/MSE/Hadamard graphs with per-output-channel UINT8 weights and scalar UINT16 activations; MinMax/MSE share all 103 weight encodings and 24 added Value grids, while 28 ordinary activation Q nodes differ. Observer-by-VRQ contrasts use 2,000 paired label-stratified draws (seed 20261006) and equal category weights.

The IQ9075 KL observer, denoted KL (2048/128), histograms every observed element into 2,048 symmetric bins and searches thresholds over 128 reconstruction bins; this is a calibration parameter, not the deployed precision (UINT16). Corrected and uncorrected KL graphs differ only by the 24 added Value QDQ pairs. Equivalence to the historical VENTUNO Q Entropy observer is not established.

\subsection{Full-dataset MSE inference on three QNN boards}
\label{app:iq_full_mse}
With calibration set B, the two frozen MSE contexts run on all 4,971 MVTec AD, VisA, and AD2 public targets. Immutable packs, cached reload, CPU-fallback disabling, and hash checks cover all 70 category--model outputs; official-evaluator spot checks differ by at most $2.473\times10^{-9}$. Each method has one cached-context session per dataset with two warm-ups; Table~\ref{tab:multidata} and Table~\ref{tab:devices} report quality and p50.

VENTUNO Q and RB5 Gen2 repeat the same graphs and targets with the same runtime versions/checks; RB5 uses an archived FastRPC discovery adapter and stock QNN libraries for inference.

\subsection{Quantization units and MSE implementation}
\label{app:mse_implementation}
Follow-up graphs use UINT8 weights per output channel and UINT16 activations per tensor: six Conv weights (axis 0), 96 attention/multilayer-perceptron Gemm projections (\texttt{transB=0}, axis 1), and one image-predictor Gemm (\texttt{transB=1}, axis 0). Added Value encodings use one scalar scale and zero point per block, not per head, token or channel.

MSE denotes a fixed reconstruction-error range search, not every MSE observer. For each FP tensor, it keeps up to 256 values per source image plus both extrema, enumerates lower/upper tail fractions from $\{0,10^{-4},10^{-3},0.005,0.02\}$ (25 ranges, all containing zero), and selects the affine UINT16 range with the lowest mean squared reconstruction error; labels are not used, and layout-range propagation is retained before the added V boundary. Follow-up exports have 809 ordinary activation Q nodes, whereas set A uses the original export with 808 nodes, changing only the objective and copying set A Value grids.

\subsection{MinMax and MSE without and with VRQ}
For each observer, VRQ Off removes only the 24 added V QDQ pairs and their 48 scalar initializers, rewiring PV to the preceding dequantize output; restoring them reconstructs the original ONNX bytes, while V keeps its inherited activation quantization. Table~\ref{tab:bottle_observer} merges observer and rotation contrasts on the same images.

\begin{table*}[!t]
\centering\small
\caption{Joint IQ9075 observer and Hadamard ablations: six unique conditions using the same VisA 256 calibration images, per-output-channel weights and 83 bottle targets. Mean is $(\mathrm{Image}+\mathrm{Pixel96\text{-}NN})/2$; bold marks the highest mean among these six conditions. Shared MinMax predictions are bit-identical across the two campaigns and shown once. Timing remains in separate columns (median of two session p50 values, ms); a dash means that condition was not timed in that campaign. The folded $H_{64}$ rows use MinMax ordinary calibration.}
\label{tab:bottle_observer}
\label{tab:bottle_rotation}
\setlength{\tabcolsep}{5pt}
\begin{adjustbox}{max width=\linewidth}
\begin{tabular}{lllrrrrr}
\toprule
 &  &  &  &  &  & \multicolumn{2}{c}{\textbf{Model p50 by campaign (ms)}}\\
\cmidrule(lr){7-8}
\textbf{Observer} & \textbf{Rotation} & \textbf{VRQ} & \textbf{Image} & \textbf{Pixel96-NN} & \textbf{Mean} & \textbf{Observer} & \textbf{Rotation}\\
\midrule
MinMax & Identity & Off & 97.7778 & 93.4127 & 95.5952 & 610.12 & 611.81 \\
\textbf{MinMax} & \textbf{Identity} & \textbf{On} & 98.4127 & 93.4466 & \textbf{95.9297} & 649.89 & 646.28 \\
MSE & Identity & Off & 98.1746 & 93.4771 & 95.8258 & 608.39 & --- \\
MSE & Identity & On & 98.3333 & 93.4440 & 95.8887 & 646.92 & --- \\
MinMax & Fixed $H_{64}$ & Off & 98.0952 & 93.5232 & 95.8092 & --- & 611.00 \\
MinMax & Fixed $H_{64}$ & On & 98.2540 & 93.4735 & 95.8637 & --- & 646.64 \\
 
\bottomrule
\end{tabular}
\end{adjustbox}
\end{table*}

\subsection{Detailed observer contrasts across source calibration sets}
\label{app:observer_contrasts}
Adding V-MinMax changes MinMax Image/Pixel96-NN AUROC by +0.6349/+0.0339\,pp and MSE by +0.1587/$-0.0330$\,pp, at +39.78\,ms (+6.52\%) and +38.53\,ms (+6.33\%) model p50. Without the boundary, MSE exceeds MinMax on both metrics (Table~\ref{tab:bottle_observer}); with it, Image ordering reverses (one net pair). One reversal among the 1,260 normal--anomalous pairs is 0.07937\,pp, and V-MinMax gains eight net pairs for MinMax and two for MSE. A 5,000-draw class-stratified paired bootstrap (seed 9075) gives Image-gain 95\% intervals $[-0.2381,2.1429]$ and $[-0.4762,0.8730]$\,pp and a gain difference of +0.4762\,pp $[-0.1587,1.6667]$, so a general observer interaction is not established. One-session MSE checks give VRQ off/on p50 of 604.85/644.15\,ms (Real-IAD bottle) and 604.26/642.60\,ms (Real-IAD toothbrush); VisA-calibrated toothbrush values appear above.

\paragraph{Toothbrush and calibration set B}
On toothbrush (12/30), MinMax PTQ+VRQ and uncorrected MSE tie at 91.9444 Image AUROC, versus 91.3889 for MSE PTQ+VRQ; rotated MinMax PTQ+VRQ has the highest Pixel AUROC (92.7451) and AUPRO (70.8290 at 0.05, 91.6035 at 0.30). With calibration set B, MSE again starts above MinMax on bottle (98.4921 versus 98.1746), and VRQ reverses the ordering; on toothbrush, both reach 91.9444 with VRQ. The two-category VRQ Image gain is 0.8036\,pp larger for MinMax than MSE ($[-0.1389,2.3021]$\,pp), yet uncorrected MSE has the highest two-category mean (95.4266). KL (2048/128) shows further category dependence: VRQ raises bottle Image AUROC by 2.2619\,pp but lowers toothbrush by 4.7222\,pp, while original-mask Pixel AUROC rises on both. No general observer ranking follows from these small splits.

\subsection{Hadamard comparison without and with VRQ}
The tested rotation is a fixed normalized Sylvester Hadamard matrix $H_{64}$ within each of the 16 Value heads in all 24 blocks, without sign randomization. With $R=\operatorname{diag}(H_{64},\ldots,H_{64})$ and row-vector activations, the FP transformation is
\begin{equation}
W_V'=W_VR,\quad b_V'=b_VR,\quad W_O'=R^{\mathsf T}W_O.
\end{equation}
Since $RR^{\mathsf T}=I$, $(PVR)(R^{\mathsf T}W_O)=PVW_O$; Query, Key, and operator topology are unchanged, while weights and all activation ranges, including V-MinMax, are recalibrated from the same source set. This is a local Value/output-projection experiment, not a full-model rotation benchmark.

FP parity on all 83 inputs is within $1.97\times10^{-6}$/$8.11\times10^{-6}$ for image/pixel logits. Without the added V boundary, rotation changes Image/Pixel96-NN AUROC by +0.3175/+0.1105\,pp; with it, by $-0.1587$/+0.0268\,pp, so only the Image direction reverses. V QDQ adds 34.47\,ms without and 35.64\,ms with rotation, and folding adds no operators. Original-mask Image/Pixel means give the same bottle comparison (VisA source 96.6488 versus 96.5827 with rotation; set B 96.7041 versus 96.5641). The rotation-by-VRQ Image interaction is $-0.4762$\,pp (95\% interval $[-2.1429,0.6349]$), a controlled point estimate rather than evidence that V QDQ generally harms rotation.

\subsection{Numerical fidelity and post-hoc mechanism checks}
On FP bottle tensors, the median maximum-to-mean channel-energy ratio within a Value head falls from 2.5054 to 2.1170 after rotation (270/384 heads improve), so rotation reduces measured concentration even though Image AUROC with V QDQ falls.

\begin{table}[!tbp]
\centering\small
\caption{Native logit MSE relative to the FP reference on bottle. Entries are MSE multiplied by $10^4$; lower is better. All rows have the added V-MinMax boundary.}
\label{tab:bottle_fidelity}
\setlength{\tabcolsep}{3pt}
\begin{adjustbox}{max width=\linewidth}
\begin{tabular}{lrr}
\toprule
\textbf{Condition} & \textbf{Image MSE} & \textbf{Pixel MSE}\\
\midrule
MinMax + V-MinMax & 8.1629 & 2.6955 \\
MSE + V-MinMax & 7.5959 & 2.7225 \\
MinMax + Hadamard + V-MinMax & 3.9591 & 2.8989 \\
 
\bottomrule
\end{tabular}
\end{adjustbox}
\end{table}

MSE and rotation lower image-logit reconstruction error but also lower Image AUROC relative to MinMax with V-MinMax (Table~\ref{tab:bottle_fidelity}). A two-stage QDQ simulation finds 57 source-range exceedances among about $4.72\times10^9$ Value elements before rotation and 784 after. This diagnostic does not explain the native AUROC changes.

\subsection{Changing only the Value range rule}
Keeping the MSE baseline's weights and ordinary activations, Value-MSE applies the same 25-candidate search to 65,536 sampled values per block plus both extrema. All 24 blocks select the MinMax range, so ONNX bytes, cached context, and outputs are unchanged; the p50 difference in Table~\ref{tab:bottle_v_rules} reflects repeated execution of one model.

\begin{table}[!tbp]
\centering\small
\caption{Value-only range rules with MSE ordinary activations and fixed per-channel weights. Each block is a separate two-session-per-arm campaign. The two MSE-search rows are byte-identical models.}
\label{tab:bottle_v_rules}
\setlength{\tabcolsep}{2.6pt}
\begin{adjustbox}{max width=\linewidth}
\begin{tabular}{llrrr}
\toprule
\textbf{Campaign} & \textbf{V rule} & \textbf{Image} & \textbf{Pixel96-NN} & \textbf{p50 (ms)}\\
\midrule
MSE search & MinMax & 98.3333 & 93.4440 & 645.68 \\
MSE search & MSE & 98.3333 & 93.4440 & 646.26 \\
\midrule
Percentile & MinMax & 98.3333 & 93.4440 & 646.67 \\
Percentile & Central 99.99\% & 98.3333 & 93.4621 & 645.42 \\
 
\bottomrule
\end{tabular}
\end{adjustbox}
\end{table}

The percentile rule uses quantiles 0.00005 and 0.99995 of the same cache without extrema; all 24 grids change (median scale ratio 0.5644). Pixel96-NN improves by +0.0181\,pp, Image AUROC is unchanged (one pair gained, one lost), and p50 is lower in Table~\ref{tab:bottle_v_rules}; neither better generalization nor faster execution is established.

\subsection{Supplementary weight-granularity control}
Table~\ref{tab:bottle_initial} varies weight granularity or ordinary observer while keeping source-MinMax Value grids. On the same Gemm export, per-channel weights improve both metrics at higher observed latency; this is not a universal granularity ranking.

\begin{table}[!tbp]
\centering\small
\caption{Initial IQ9075 bottle controls. Every row includes V-MinMax; one timed session per row. W denotes weight granularity.}
\label{tab:bottle_initial}
\setlength{\tabcolsep}{3pt}
\begin{adjustbox}{max width=\linewidth}
\begin{tabular}{lrrr}
\toprule
\textbf{Ordinary observer / W} & \textbf{Image} & \textbf{Pixel96-NN} & \textbf{p50 (ms)}\\
\midrule
MinMax, per-tensor W & 98.1349 & 93.3922 & 626.48 \\
MinMax, per-channel W & 98.4127 & 93.4466 & 650.47 \\
MSE, per-channel W & 98.3333 & 93.4440 & 647.24 \\
 
\bottomrule
\end{tabular}
\end{adjustbox}
\end{table}

\subsection{Reproducibility}
Reproducibility records include input manifests, model and runtime identifiers, graph checks, range-search code and session timings.

Figure~\ref{fig:bottle_interactions} motivates joint native evaluation of activation calibration, weight granularity and Value boundaries; it establishes no universal observer, hardware independence or unseen-category performance, and QDQ metadata need not match compiled precision.

\section{Paired Aggregate Uncertainty and Sustained Execution}
\label{app:completed_evidence}
\setcounter{table}{0}
\renewcommand{\thetable}{\Alph{section}\arabic{table}}
\renewcommand{\theHtable}{\Alph{section}\arabic{table}}

\subsection{Paired cluster uncertainty on saved outputs}
This analysis reuses retained native outputs of the six matched set A configurations, with no new calibration or model selection; recomputed category AUROCs match the retained scores. A cluster is one image for MVTec AD and VisA and one physical scene with all lighting views for AD2 public (1,725, 2,162 and 184 clusters). A paired deletion removes the same cluster from both configurations and both Image/Pixel statistics, preserving within-image pixel dependence; pixel scores keep exact float32 ordering with half credit for ties. A synthetic check of 720 deletion cases agrees with direct recomputation to $1.12\times10^{-16}$.

Let $\widehat\delta_c$ denote a paired difference in category $c$, $n_c$ its cluster count, and $\widehat\delta_{c,(-j)}$ the difference after deleting cluster $j$. The category-level variance estimate is
\begin{equation}
\widehat v_c=\frac{n_c-1}{n_c}\sum_{j=1}^{n_c}
\left(\widehat\delta_{c,(-j)}-\overline{\delta}_{c,(-\cdot)}\right)^2.
\end{equation}
For $C_d$ fixed categories in dataset $d$, the macro variance is $\sum_c\widehat v_c/C_d^2$, and the three-dataset variance is the sum of the three macro variances divided by nine; Image/Pixel means are formed within every deletion to retain covariance. Intervals are estimate $\pm1.96$ standard errors: post-hoc, descriptive normal approximations conditional on fixed calibration, categories and export, without multiplicity correction.

Table~\ref{tab:cluster_recovery} lists all seven specified contrasts. Source-MinMax Value correction improves the three-dataset AUROC mean under both ordinary observers, through tensor overrides or added QDQ. Contrasts between the two implementations include zero, establishing neither superiority nor equivalence; Pixel-only intervals versus each PTQ baseline also include zero.

\begin{table*}[t]
\centering\small
\caption{Calibration set A paired AUROC differences (pp), averaged equally across the three datasets. Each metric is shown as its difference and a paired cluster-jackknife 95\% interval. The last pair averages Image and Pixel equally. Override uses the matched Value-MinMax grid; intervals are conditional and multiplicity-unadjusted. All seven contrasts are retained.}
\label{tab:cluster_recovery}
\setlength{\tabcolsep}{3pt}
\begin{adjustbox}{max width=\linewidth}
\begin{tabular}{lrrrrrr}
\toprule
 & \multicolumn{2}{c}{\textbf{Image AUROC}} & \multicolumn{2}{c}{\textbf{Pixel AUROC}} & \multicolumn{2}{c}{\textbf{3-set AUROC Mean}}\\
\textbf{Contrast} & $\boldsymbol\Delta$ & \textbf{95\% interval} & $\boldsymbol\Delta$ & \textbf{95\% interval} & $\boldsymbol\Delta$ & \textbf{95\% interval}\\
\midrule
MinMax: V (override) $-$ PTQ & +1.670 & [+0.807, +2.534] & +0.040 & [-0.174, +0.255] & +0.855 & [+0.404, +1.307]\\
MinMax: VRQ $-$ PTQ & +1.756 & [+0.902, +2.610] & +0.048 & [-0.164, +0.260] & +0.902 & [+0.456, +1.348]\\
MinMax: VRQ $-$ V (override) & +0.085 & [-0.205, +0.375] & +0.008 & [-0.003, +0.019] & +0.047 & [-0.098, +0.192]\\
MSE: V (override) $-$ PTQ & +1.363 & [+0.577, +2.148] & +0.032 & [-0.198, +0.261] & +0.697 & [+0.283, +1.111]\\
MSE: VRQ $-$ PTQ & +1.285 & [+0.497, +2.073] & +0.033 & [-0.192, +0.259] & +0.659 & [+0.244, +1.074]\\
MSE: VRQ $-$ V (override) & -0.078 & [-0.359, +0.203] & +0.002 & [-0.007, +0.011] & -0.038 & [-0.179, +0.103]\\
MinMax PTQ+VRQ $-$ MSE PTQ+VRQ (override) & +0.033 & [-0.257, +0.322] & +0.013 & [+0.000, +0.026] & +0.023 & [-0.122, +0.168]\\
 
\bottomrule
\end{tabular}
\end{adjustbox}
\end{table*}

Full-precision retained scores differ from rounded set A cells of Table~\ref{tab:multidata} by at most 0.001 in displayed means. All 210 category--configuration deletion records are retained, and independent aggregation reproduces the summary within $10^{-12}$\,pp.

\subsection{Ten-minute cached-context workloads}
Each condition runs a frozen, hash-verified VENTUNO Q context for at least 600\,s, cycling the first 64 images of the set A reserve partition, preprocessed before timing. Each process checks three compile-time sentinel outputs and performs three warm-ups; ONNX Runtime 1.30.0 and QNN EP plugin 2.6.0 use recorded HTP options with CPU fallback disabled, and every profile contains only QNN node execution. The competing application is stopped during measurement and restored afterward; thermal state, host resident memory and load are sampled every 16 calls, and repeated outputs are compared with first-cycle hashes. Model p50/p95 excludes loading, warm-up, preprocessing and telemetry; peak host resident memory covers the whole worker lifetime, not total NPU memory. Seed 20261109 fixes the order; interrupted attempts are archived separately and not pooled, and the four remaining conditions resumed with identical worker, protocol, input and runtime hashes. Table~\ref{tab:sustained_recovery} reports six complete runs (8,950 measured calls), all bit-exact within condition; 60-second windows characterize one run, not independent repetitions.

\begin{table*}[t]
\centering\small
\caption{VENTUNO Q sustained model workload: one complete 600-second run per condition on the same cyclic 64 reserve inputs. Timing is in ms; peak host-process RSS is in MiB. The window column gives the minimum and maximum p50 across ten 60-second windows. These measurements are separate from Table~\ref{tab:devices}'s target and short-session timing campaigns.}
\label{tab:sustained_recovery}
\setlength{\tabcolsep}{4pt}
\begin{adjustbox}{max width=\linewidth}
\begin{tabular}{lrrrrr}
\toprule
\textbf{Condition} & \textbf{Calls} & \textbf{Model p50} & \textbf{Model p95} & \textbf{Window p50 range} & \textbf{Peak host RSS}\\
\midrule
MinMax PTQ & 1525 & 385.27 & 388.90 & 385.08--387.92 & 956.12\\
MinMax PTQ + VRQ (override) & 1481 & 397.91 & 401.60 & 397.67--398.13 & 956.01\\
MinMax PTQ + VRQ & 1470 & 400.37 & 404.03 & 400.10--400.67 & 956.11\\
MSE PTQ & 1522 & 386.25 & 389.87 & 386.01--389.01 & 956.09\\
MSE PTQ + VRQ (override) & 1470 & 400.59 & 404.29 & 400.39--401.42 & 956.15\\
MSE PTQ + VRQ & 1482 & 397.17 & 400.86 & 396.94--397.58 & 956.13\\
 
\bottomrule
\end{tabular}
\end{adjustbox}
\end{table*}

\section{Additional analyses and alternative protocols}
\label{app:secondary}
\setcounter{table}{0}
\renewcommand{\thetable}{\Alph{section}\arabic{table}}
\renewcommand{\theHtable}{\Alph{section}\arabic{table}}
\subsection{Observer and rotation interactions}
\label{sec:vrq_observer_story}
\label{sec:observer_rotation_interactions}
Figure~\ref{fig:bottle_interactions} summarizes the exploratory IQ9075 ablations detailed in Appendix~\ref{app:observer_contrasts}. MinMax and MSE share the corrected Value grids but retain different upstream encodings. The folded $H_{64}$ control tests a local rotation inspired by QuaRot, SpinQuant and QuantVGGT~\cite{quarot,spinquant,quantvggt}, rather than those full methods. With VRQ, MSE and rotation both reduce logit reconstruction error relative to MinMax+VRQ but have lower Image AUROC point estimates; the interaction intervals include zero. These controls distinguish tensor fidelity from anomaly-score ordering (Appendix~\ref{app:bottle_followup}).
\begin{figure*}[t]
\centering
\includegraphics[width=\textwidth]{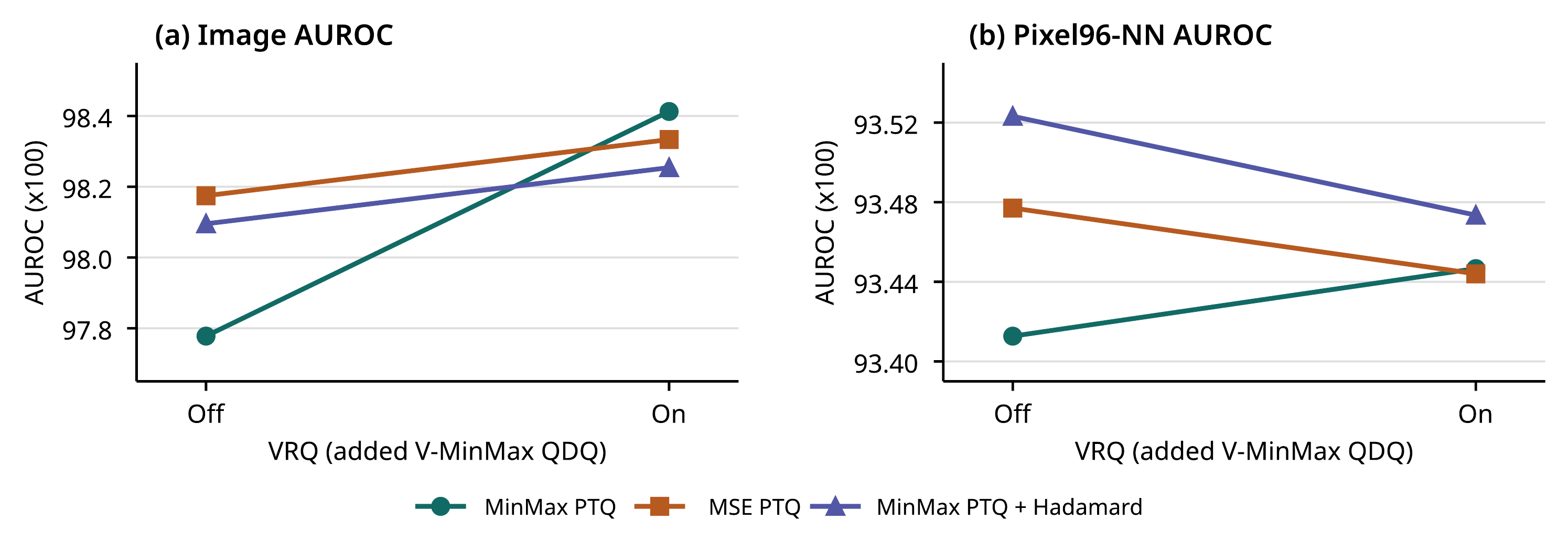}
\caption{IQ9075 bottle observer/rotation interactions with the same VisA calibration list and per-channel weights. Lines join VRQ off/on point estimates; Pixel96-NN uses nearest-resized masks. Panels use different vertical scales. Quality predictions are identical across timing campaigns; timings are not pooled. Paired Image intervals are reported in Appendix~\ref{app:observer_contrasts}.}
\label{fig:bottle_interactions}
\end{figure*}

\subsection{Observer-dependent transfer}
On RB5 Gen2 (Table~\ref{tab:controls}), Entropy+\vr{} raises Image AUROC from 96.326 to 96.523 (+\EntropyImageDelta{}\,pp), with a \nativepixel{} gain of \EntropyPixelDelta{}\,pp. Percentile+\vr{} raises Image AUROC from 72.437 to 73.805 but lowers \nativepixel{} from 91.356 to 91.255: correcting V leaves other quantization errors intact.

\subsection{Category variation and qualitative decisions}
\paragraph{Lighting-dependent failure}
Although the AD2 mean improves, Entropy PTQ+\vr{} lowers \texttt{rice} Image AUROC by $-7.116$\,pp (scene-cluster 95\% interval [$-14.221$, $-0.450$]\,pp), with a decrease under every lighting condition (Appendix~\ref{app:lighting}), while its original-mask Pixel AUROC rises by +2.648\,pp. Localization-ranking recovery does not ensure image-ranking recovery.

\begin{figure*}[t]
\centering
\includegraphics[width=.94\textwidth]{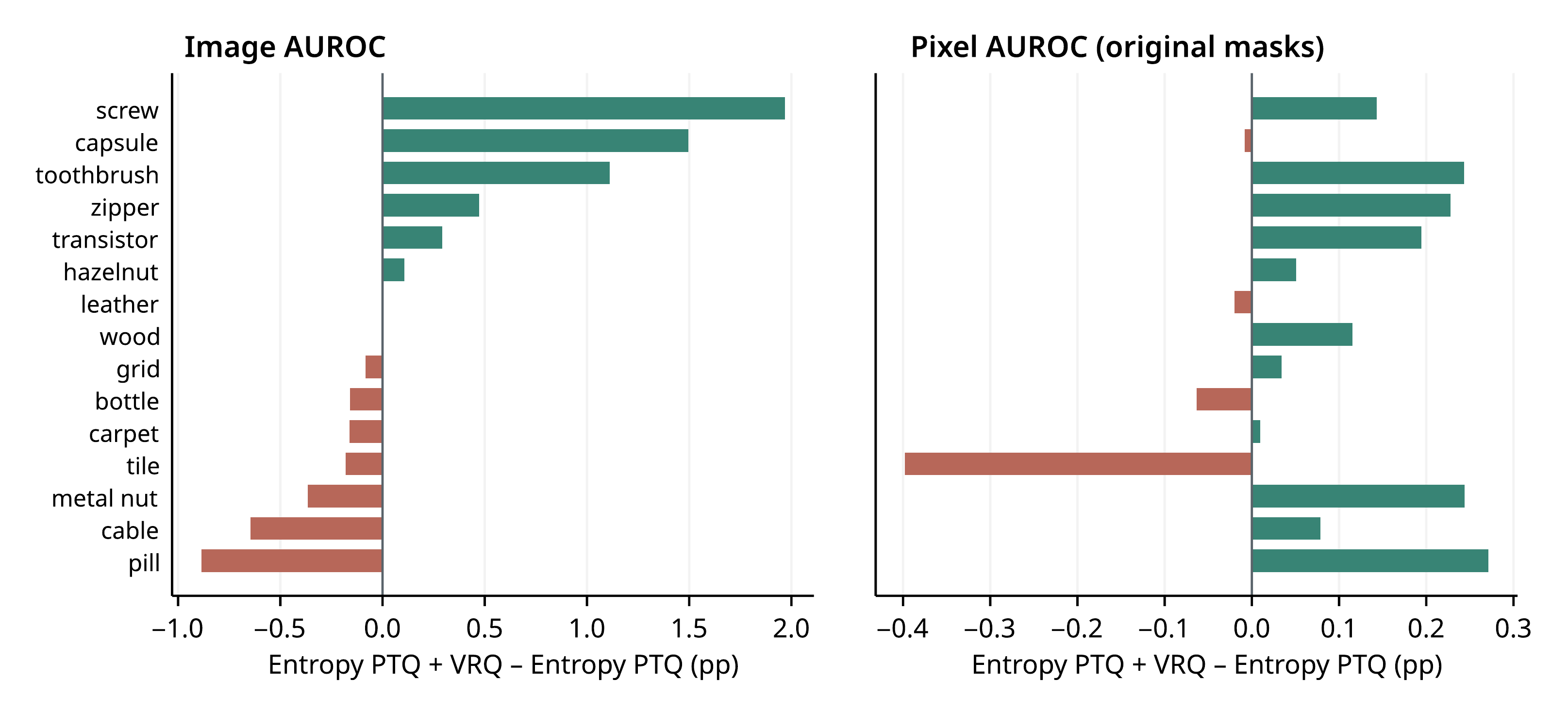}
\caption{Entropy PTQ+\vr{} minus Entropy PTQ on VENTUNO Q (set A, pp), across all 15 MVTec AD categories ordered by Image gain. Pixel AUROC uses original masks and resized predictions. Green/red denotes increase/decrease; panels use different axis ranges.}
\label{fig:categories}
\end{figure*}

\paragraph{Category-level heterogeneity}
Across the 15 MVTec AD categories in Figure~\ref{fig:categories}, Image AUROC improves in six, is unchanged in two, and decreases in seven; original-mask Pixel AUROC improves in eleven and decreases in four. Macro gains are +0.198/+0.075\,pp. Image changes range from +1.968\,pp (screw) to $-0.887$\,pp (pill). Cable loses $-0.647$\,pp in Image AUROC while gaining +0.078\,pp in Pixel AUROC, illustrating the distinct ranking tasks.

\begin{figure*}[!tp]
\centering
\includegraphics[width=.94\textwidth]{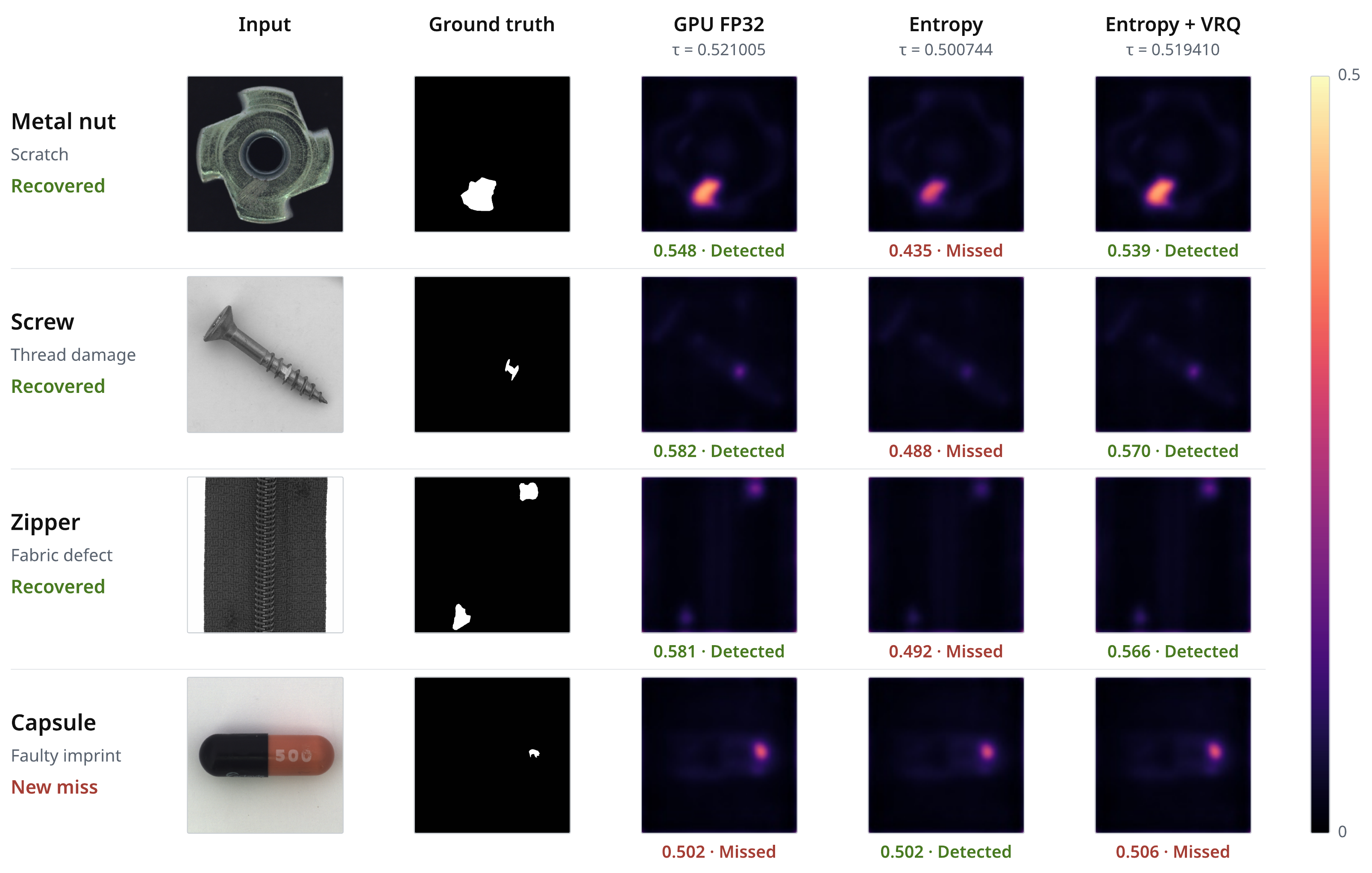}
\caption{Three recoveries and one new miss, selected from 36 recovered and 7 newly missed MVTec anomalies. Native $96\times96$ maps share a linear 0--0.5 scale; decisions use strict $>$ on unrounded scores at fixed source thresholds.}
\label{fig:qualitative}
\end{figure*}

\paragraph{Qualitative operating-point examples} Figure~\ref{fig:qualitative} illustrates the recoveries and losses in Table~\ref{tab:retention}. Three post-hoc selections show recovered metal-nut scratches, screw-thread damage, and zipper-fabric defects; the capsule is the first lexicographic new miss. These illustrate outcomes rather than representative frequencies. The capsule is missed despite a higher score because its source-fixed threshold also rises; FP32 misses it too. Image decisions and localization appearance are distinct outputs.

\subsection{Complete quality comparisons and alternative calibration implementations}
\label{app:complete_quality}
Table~\ref{tab:complete_auroc} gives the full AUROC comparison; Table~\ref{tab:complete_aupro} adds the set B controls omitted from the main AUPRO table. Set B's KL observer differs from set A's Entropy implementation; the cause of its lower quality, and of the Percentile control in Table~\ref{tab:controls}, remains unresolved.
\begin{table*}[!tp]
\centering\small\renewcommand{\arraystretch}{.95}
\caption{Complete Image/original-mask Pixel AUROC ($\times100$) on 4,971 targets. Sets A/B retain their respective source inputs, exports and VENTUNO Q/IQ9075 runtimes (Table~\ref{tab:multidata}); VRQ (override) uses the matched Value grid. Dataset means average Image and Pixel; the 3-set mean weights datasets equally. Bold ranks descriptive means among quantized conditions within each set. A uses three-decimal cells; B uses unrounded scores.}
\label{tab:complete_auroc}

\setlength{\tabcolsep}{2.3pt}
\begin{adjustbox}{max width=\linewidth}
\begin{tabular}{llrrrrrrrrrr}
\toprule
 &  & \multicolumn{3}{c}{\textbf{MVTec AD (1,725)}} & \multicolumn{3}{c}{\textbf{VisA (2,162)}} & \multicolumn{3}{c}{\textbf{AD2 public (1,084)}} & \textbf{3-set}\\
\cmidrule(lr){3-5}\cmidrule(lr){6-8}\cmidrule(lr){9-11}
\textbf{Condition} & \textbf{Calibration set} & \textbf{Image} & \textbf{Pixel} & \textbf{Mean} & \textbf{Image} & \textbf{Pixel} & \textbf{Mean} & \textbf{Image} & \textbf{Pixel} & \textbf{Mean} & \textbf{AUROC Mean}\\
\midrule
GPU FP32 & A & 96.445 & 92.050 & 94.248 & 92.888 & 95.630 & 94.259 & 67.705 & 87.230 & 77.468 & 88.658 \\
HTP FP16 & A & 96.448 & 92.057 & 94.253 & 92.932 & 95.637 & 94.285 & 67.650 & 87.252 & 77.451 & 88.663 \\
\midrule
MinMax PTQ & A & 96.047 & 91.990 & 94.019 & 91.565 & 95.503 & 93.534 & 64.090 & 87.282 & 75.686 & 87.746 \\
MinMax PTQ & B & 96.236 & 91.973 & 94.105 & 91.482 & 95.477 & 93.480 & 64.725 & 87.360 & 76.043 & 87.876 \\
MinMax PTQ + VRQ (override) & A & 96.463 & 92.062 & 94.263 & 92.862 & 95.608 & 94.235 & 67.389 & 87.225 & 77.307 & 88.602 \\
\textbf{MinMax PTQ + VRQ} & A & 96.529 & 92.069 & 94.299 & 93.001 & 95.609 & 94.305 & 67.440 & 87.241 & \textbf{77.341} & \textbf{88.648} \\
\textbf{MinMax PTQ + VRQ} & B & 96.463 & 92.065 & \textbf{94.264} & 92.876 & 95.623 & \textbf{94.249} & 67.428 & 87.229 & \textbf{77.329} & \textbf{88.614} \\
\midrule
Entropy PTQ & A & 96.326 & 92.004 & 94.165 & 92.118 & 95.523 & 93.821 & 65.395 & 87.377 & 76.386 & 88.124 \\
Entropy (KL) PTQ & B & 69.075 & 91.099 & 80.087 & 55.597 & 93.945 & 74.771 & 52.516 & 82.179 & 67.347 & 74.068 \\
\textbf{Entropy PTQ + VRQ} & A & 96.523 & 92.079 & \textbf{94.301} & 92.786 & 95.595 & 94.191 & 65.913 & 87.293 & 76.603 & 88.365 \\
Entropy (KL) PTQ + VRQ & B & 70.112 & 91.047 & 80.579 & 58.721 & 94.017 & 76.369 & 52.359 & 82.289 & 67.324 & 74.758 \\
\midrule
MSE PTQ & A & 96.232 & 91.994 & 94.113 & 91.553 & 95.501 & 93.527 & 64.999 & 87.289 & 76.144 & 87.928 \\
MSE PTQ & B & 96.378 & 91.995 & 94.187 & 91.687 & 95.513 & 93.600 & 65.280 & 87.374 & 76.327 & 88.038 \\
\textbf{MSE PTQ + VRQ (override)} & A & 96.412 & 92.064 & 94.238 & 93.024 & 95.607 & \textbf{94.316} & 67.434 & 87.208 & 77.321 & 88.625 \\
MSE PTQ + VRQ & A & 96.516 & 92.066 & 94.291 & 92.858 & 95.607 & 94.233 & 67.263 & 87.212 & 77.238 & 88.587 \\
MSE PTQ + VRQ & B & 96.422 & 92.065 & 94.244 & 92.832 & 95.617 & 94.225 & 67.171 & 87.249 & 77.210 & 88.560 \\
\midrule
AdaRound (fixed QNN grids) & A & 96.443 & 92.033 & 94.238 & 91.949 & 95.508 & 93.729 & 64.876 & 87.333 & 76.105 & 88.024 \\
AdaRound (fixed KL grids) & B & 67.069 & 91.122 & 79.095 & 55.438 & 94.074 & 74.756 & 52.509 & 82.184 & 67.347 & 73.733 \\
 
\bottomrule
\end{tabular}
\end{adjustbox}
\par\vspace{6pt}
\centering\small
\caption{Additional set B controls: original-mask AUPRO ($\times100$). See Table~\ref{tab:aupro} for the other conditions and the definition of the 3-set mean.}
\label{tab:complete_aupro}
\setlength{\tabcolsep}{4pt}
\begin{adjustbox}{max width=\linewidth}
\begin{tabular}{llrrrrrrr}
\toprule
 &  & \multicolumn{2}{c}{\textbf{MVTec AD (1,725)}} & \multicolumn{2}{c}{\textbf{VisA (2,162)}} & \multicolumn{2}{c}{\textbf{AD2 public (1,084)}} & \textbf{3-set}\\
\textbf{Condition} & \textbf{Calibration set} & \textbf{@0.30} & \textbf{@0.05} & \textbf{@0.30} & \textbf{@0.05} & \textbf{@0.30} & \textbf{@0.05} & \textbf{AUPRO Mean}\\
\midrule
Entropy (KL) PTQ & B & 84.251 & 60.152 & 84.127 & 53.057 & 43.518 & 20.660 & 57.628 \\
Entropy (KL) PTQ + VRQ & B & 84.212 & 59.793 & 84.062 & 52.820 & 43.819 & 20.803 & 57.585 \\
AdaRound (fixed KL grids) & B & 84.284 & 60.208 & 84.581 & 53.734 & 43.594 & 20.613 & 57.836 \\
 
\bottomrule
\end{tabular}
\end{adjustbox}
\end{table*}
\subsection{Complete device records and workload definitions}
\label{app:complete_native}
Table~\ref{tab:complete_devices} records additional device quality and timing. Table~\ref{tab:devices} reports the matched VENTUNO Q campaign. Earlier set A runs use sequential conditions except for the alternating MSE pair. Source64 uses session medians; full-target timings pool calls by dataset.
For MVTec multi-board verification and calibration repeats, \nativepixel{} resizes ground truth bilinearly to $96\times96$ and binarizes it above 0.5, whereas primary localization uses original masks; the two protocols are labeled separately. Calibration set B outputs are rescored with the original-mask metrics (Tables~\ref{tab:multidata} and~\ref{tab:aupro}); Appendix~\ref{app:bottle_followup} keeps the bottle Pixel96-NN definition.

\begin{table*}[!tp]
\centering\small\renewcommand{\arraystretch}{.9}
\caption{Additional native quality and model p50 (ms). Bold marks the highest quantized mean AUROC per board and the source64/set B timing minima. Earlier set A times are unranked; Table~\ref{tab:devices} reports the matched campaign. Appendix~\ref{app:timing_checks} defines the timing protocols.}
\label{tab:complete_devices}

\setlength{\tabcolsep}{3pt}
\noindent\begin{minipage}[t]{.48\linewidth}\vspace{0pt}\centering
\begin{adjustbox}{max width=\linewidth}
\begin{tabular}{llrrr}
\toprule
\multicolumn{5}{l}{\textbf{\textit{MVTec quality and source64 timing}}}\\
\textbf{Device} & \textbf{Condition} & \textbf{Image} & \textbf{Pixel$_{96}$} & \textbf{p50}\\
\midrule
RB5 Gen2 & HTP FP16 & 96.455 & 92.058 & 1040.58 \\
RB5 Gen2 & MinMax PTQ & 96.047 & 91.992 & \textbf{545.34} \\
RB5 Gen2 & \textbf{MinMax PTQ + \vr{}} & \textbf{96.529} & \textbf{92.070} & 559.35 \\
IQ9075 & HTP FP16 & 96.454 & 92.058 & 797.91 \\
IQ9075 & MinMax PTQ & 96.047 & 91.992 & \textbf{364.31} \\
IQ9075 & \textbf{MinMax PTQ + \vr{}} & \textbf{96.529} & \textbf{92.070} & 376.75 \\
Orin NX & Mixed FP16/FP32 & 96.427 & 92.051 & 256.47 \\
 
\bottomrule
\end{tabular}
\end{adjustbox}
\end{minipage}\hfill\begin{minipage}[t]{.50\linewidth}\vspace{0pt}\centering
\begin{adjustbox}{max width=\linewidth}
\begin{tabular}{llrrr}
\toprule
\multicolumn{5}{l}{\textbf{\textit{Set B: matched full-target p50}}}\\
\textbf{Condition} & \textbf{Device} & \textbf{MVTec} & \textbf{VisA} & \textbf{AD2}\\
\midrule
MinMax PTQ & IQ9075 & \textbf{603.66} & 603.68 & \textbf{604.18} \\
MinMax PTQ + VRQ & IQ9075 & 641.03 & 640.45 & 640.58 \\
\midrule
Entropy (KL) PTQ & IQ9075 & 609.18 & 609.16 & 609.55 \\
Entropy (KL) PTQ + VRQ & IQ9075 & 645.23 & 645.35 & 645.72 \\
\midrule
MSE PTQ & IQ9075 & 604.36 & \textbf{603.51} & 604.45 \\
MSE PTQ & VENTUNO Q & \textbf{770.90} & \textbf{777.50} & \textbf{774.60} \\
MSE PTQ & RB5 Gen2 & \textbf{427.17} & \textbf{428.78} & \textbf{419.89} \\
MSE PTQ + VRQ & IQ9075 & 641.79 & 640.68 & 641.21 \\
MSE PTQ + VRQ & VENTUNO Q & 805.23 & 810.04 & 802.12 \\
MSE PTQ + VRQ & RB5 Gen2 & 459.38 & 460.23 & 453.93 \\
\midrule
AdaRound (fixed KL grids) & IQ9075 & 610.91 & 610.22 & 610.19 \\
 
\bottomrule
\end{tabular}
\end{adjustbox}
\end{minipage}\par\vspace{5pt}\centering
\begin{adjustbox}{max width=\linewidth}
\begin{tabular}{lrrr}
\toprule
\multicolumn{4}{l}{\textbf{\textit{Set A: earlier VENTUNO Q full-target campaigns}}}\\
\textbf{Condition} & \textbf{MVTec (1,725)} & \textbf{VisA (2,162)} & \textbf{AD2 (1,084)}\\
\midrule
HTP FP16 & 863.26 & 863.54 & 863.78 \\
MinMax PTQ & 391.08 & 391.02 & 391.05 \\
MinMax PTQ + VRQ & 403.82 & 403.85 & 403.32 \\
Entropy PTQ & 391.42 & 391.87 & 391.54 \\
Entropy PTQ + VRQ & 405.14 & 405.75 & 406.07 \\
MSE PTQ & 386.19 & 387.11 & 386.52 \\
MSE PTQ + VRQ & 398.68 & 400.38 & 399.74 \\
AdaRound (fixed grids) & 391.69 & 392.07 & 392.00 \\
 
\bottomrule
\end{tabular}
\end{adjustbox}
\end{table*}

\clearpage
\end{document}